\documentclass[10pt,twocolumn,letterpaper]{article}

\usepackage[pagenumbers]{conference} %

\definecolor{cvprblue}{rgb}{0.21,0.49,0.74}
\usepackage[pagebackref,breaklinks,colorlinks,allcolors=cvprblue]{hyperref}

\usepackage{relsize}
\usepackage{tabularx}
\usepackage{array}
\usepackage[accsupp]{axessibility}  %
\newsavebox\CBox

\usepackage{xcolor}
\usepackage{overpic}
\usepackage{colortbl}

\usepackage{graphicx}

\usepackage{caption}

\graphicspath{{images/}}
\newcommand{\tabref}[1]{Tab.~\ref{#1}}

\newcommand{\equref}[1]{Equ.~\ref{#1}}
\newcommand{\figrefn}[1]{Fig.~\ref{#1}}
\newcommand{\secrefn}[1]{Sec.~\ref{#1}}
\usepackage{amsmath}
\usepackage{dsfont}
\usepackage{graphicx} %
\usepackage{caption}  %
\usepackage{capt-of}  %
\usepackage{cuted}
\usepackage{enumitem}
\usepackage{booktabs}
\usepackage{graphicx} %
\usepackage{caption} %

\title{3DGS-DET: Empower 3D Gaussian Splatting with Boundary Guidance and Box-Focused Sampling for Indoor 3D Object Detection}

\author{%
 Yang Cao$^{*}$, Yuanliang Ju$^{*}$, Dan Xu$^{\dag}$ \vspace{0.1cm} \\
Hong Kong University of Science and Technology}

\begin{document}

\twocolumn[{%
\maketitle
\renewcommand\twocolumn[1][]{#1}%
\begin{center}
    \centering
    \captionsetup{type=figure}
    \vspace{-7mm}
\includegraphics[width=0.93\linewidth]{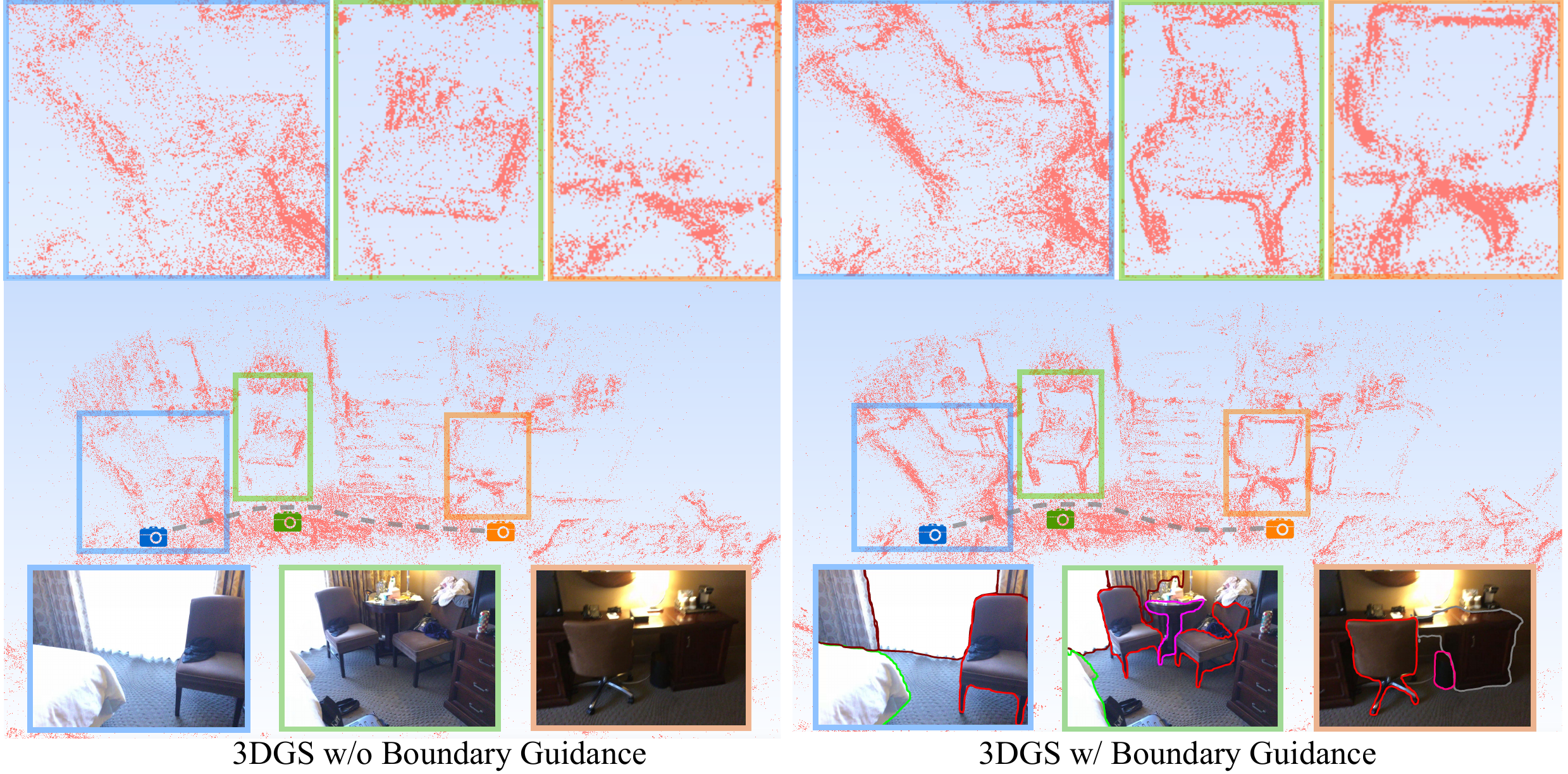}
    \captionof{figure}{Illustration of the proposed Boundary Guidance. By incorporating Boundary Guidance in the training of 3D Gaussian Splatting (3DGS), we significantly improve the spatial distribution of Gaussian blobs relating objects and the background. To better show this improved spatial distribution, we visualize only the positions of the Gaussian blobs, omitting other attributes for clarity.}
    \label{fig:teaser}
\end{center}
}]

\begingroup
\renewcommand\thefootnote{}\footnotetext{Under review.}\addtocounter{footnote}{0}
\endgroup

\begin{abstract}
Neural Radiance Fields (NeRF) have been adapted for indoor 3D Object Detection (3DOD), offering a promising approach to indoor 3DOD via view-synthesis representation. 
But its implicit nature limits representational capacity.
Recently, 3D Gaussian Splatting (3DGS) has emerged as an explicit 3D representation that addresses the limitation. This work introduces 3DGS into indoor 3DOD for the first time, identifying two main challenges: (i) Ambiguous spatial distribution of Gaussian blobs -- 3DGS primarily relies on 2D pixel-level supervision, resulting in unclear 3D spatial distribution of Gaussian blobs and poor differentiation between objects and background, which hinders indoor 3DOD; (ii) Excessive background blobs -- 2D images typically include numerous background pixels, leading to densely reconstructed 3DGS with many noisy Gaussian blobs representing the background, negatively affecting detection. To tackle (i), we leverage the fact that 3DGS reconstruction is derived from 2D images, and propose an elegant solution by incorporating 2D Boundary Guidance to significantly enhance the spatial distribution of Gaussian blobs, resulting in clearer differentiation between objects and their background (please see~\figrefn{fig:teaser}). To address (ii), we propose a Box-Focused Sampling strategy using 2D boxes to generate object probability distribution in 3D space, allowing effective probabilistic sampling in 3D to retain more object blobs and reduce noisy background blobs.~Benefiting from these innovations, 
3DGS-DET significantly outperforms the state-of-the-art NeRF-based method, NeRF-Det++, achieving improvements of \textbf{+6.0} on mAP@0.25 and \textbf{+7.8} on mAP@0.5 for the ScanNet, and the \textbf{+14.9} on mAP@0.25 for the ARKITScenes. The code and models will be made publicly available upon acceptance at: \href{https://github.com/yangcaoai/3DGS-DET}{https://github.com/yangcaoai/3DGS-DET}.

\end{abstract}
 
\section{Introduction}
\label{sec:introduction}

Indoor 3D Object Detection~(3DOD)~\cite{qi2017pointnet, votenet,nerfdetpp,xu2023nerf} is a fundamental task in computer vision, providing foundations for wide realistic application scenarios such as robotics and augmented reality, as accurate localization and classification of objects in 3D space are critical for these applications. Most existing indoor 3DOD methods~\cite{rukhovich2022imvoxelnet, rukhovich2022fcaf3d,cao2023coda,cao2024collaborative,cao2026vggtdet} explored using non-view-synthesis representations, including point clouds and multi-view images, to perform 3D object detection.~However, these approaches mainly focus on the perception perspective and lack the capability for novel view synthesis. 

Neural Radiance Fields (NeRF)~\cite{mildenhall2021nerf} provide an effective manner for novel view synthesis and have been adapted for indoor 3D Object Detection (3DOD) through view-synthesis representations~\cite{xu2023nerf, nerfdetpp}. However, as a view-synthesis representation for indoor 3DOD, NeRF has an inherent key limitation: Its implicit nature restricts its representational capacity for indoor 3DOD. Recently, 3D Gaussian Splatting (3DGS)~\cite{kerbl3Dgaussians} has emerged as an explicit 3D representation that effectively addresses the limitation. Inspired by these strengths, our work is \emph{the first} to introduce 3DGS into indoor 3DOD. In this exploration, we identify two primary challenges: 
 (i) Ambiguous spatial distribution of Gaussian blobs -- 3DGS primarily relies on 2D pixel-level supervision, resulting in unclear 3D spatial distribution of Gaussian blobs and insufficient differentiation between objects and background, which hinders effective indoor 3DOD; (ii) Excessive background blobs -- 2D images typically contain numerous background pixels, leading to densely populated 3DGS with many noisy Gaussian blobs representing the background, hindering the detection of foreground 3D objects.

\par
To address the above-discussed challenges, we further empower 3DGS with two novel strategies for 3D object detection (i)~\emph{2D Boundary Guidance Strategy}: Given the fact that 3DGS reconstruction is optimized from 2D images, we introduce a novel strategy by incorporating 2D Boundary Guidance to achieve a more suitable 3D spatial distribution of Gaussian blobs for detection. Specifically, we first perform object boundary detection on posed images, then overlay the boundaries onto the images, and finally train the 3DGS model. This proposed strategy can facilitate the learning of a spatial Gaussian blob distribution that is more differentiable for the foreground objects and the background (see~\figrefn{fig:teaser}).
(ii)~\emph{Box-Focused Sampling Strategy}:~This strategy further leverages 2D boxes to establish 3D object probability spaces, enabling an object probabilistic sampling of Gaussian blobs to effectively preserve object blobs and prune background blobs. Specifically, we project the 2D boxes that cover objects in images into 3D spaces to form frustums.~The 3D Gaussian blobs within the frustum have a higher probability of being object blobs compared to those outside. Based on this strategy, we construct 3D object probability spaces and sample Gaussian blobs accordingly, finally preserving more object blobs and reducing noisy background blobs.
In summary, the contributions of this work are fourfold:
\begin{itemize}[leftmargin=*]
\item To the best of our knowledge, we are the first to integrate 3DGS into indoor 3D Object Detection (3DOD), representing a novel contribution. We propose a novel method \textbf{3DGS-DET}, which empowers 3DGS with Boundary Guidance and Box-Focused Sampling for indoor 3DOD.
\item We design \emph{Boundary Guidance} to optimize 3DGS with the guidance of object boundaries, which achieves a significantly better spatial distribution of Gaussian blobs and clearer differentiation between objects and the background, thereby effectively enhancing indoor 3DOD.
\item We propose \emph{Box-Focused Sampling}, which establishes 3D object probability spaces, enabling a higher sampling probability to be assigned to object-related 3D Gaussian blobs.~This probabilistic sampling strategy preserves more foreground object blobs and suppresses noisy background blobs, further improving detection performance.
\item Boundary Guidance and Box-Focused Sampling improve detection by \textbf{5.6 points} on mAP@0.25 and \textbf{3.7 points} on mAP@0.5 as demonstrated in our ablation study. Furthermore, our final method, 3DGS-DET, significantly outperforms the state-of-the-art NeRF-based method, NeRF-Det++~\cite{nerfdetpp}, on both ScanNet~(\textbf{+6.0} on mAP@0.25, \textbf{+7.8} on mAP@0.5) and ARKITScenes~(\textbf{+14.9} on mAP@0.25).
Moreover, our method also clearly outperforms methods that use multi-view images as input, demonstrating the superiority of our 3DGS for indoor 3DOD.
\end{itemize}

\begin{figure*}[h!]
     \centering
    \begin{overpic}[width=0.90\textwidth]%
    {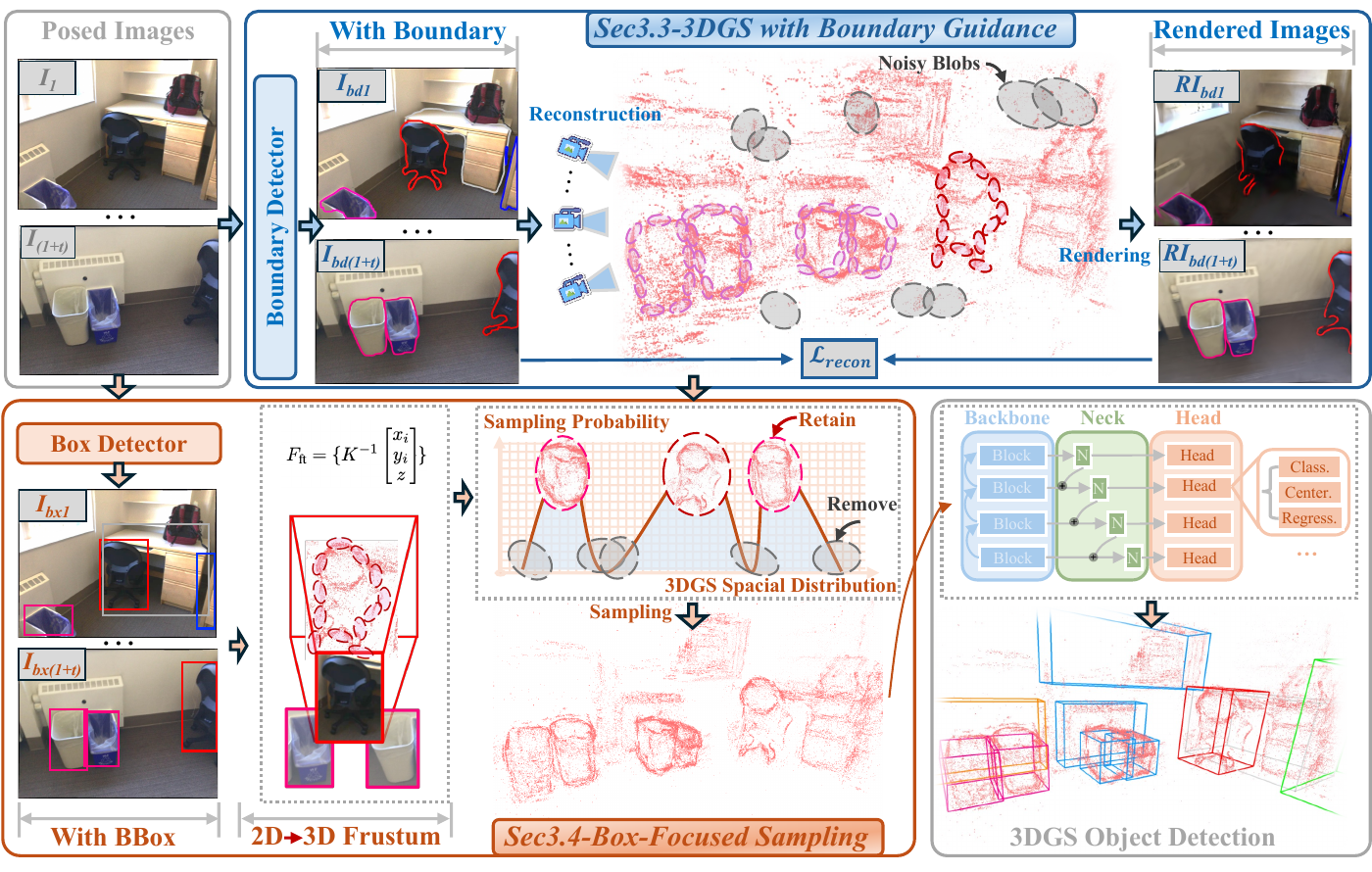}
    \end{overpic}
    \vspace{-0.3cm}  
    \caption{Method Overview (best view in color). The blue block in the top row illustrates our Boundary Guidance~(\secrefn{sec:methods:boundary-guidance}).
    The orange block in the bottom row
    shows our Box-Focused Sampling~(\secrefn{sec:methods:box-focused-sampling}). 
First, we train the 3DGS with the Boundary Guidance as shown in the blue block,
which improves the 3D spatial distribution of Gaussian blobs, and thus produces clearer differentiation between objects and the background, as highlighted by the colorful dashed ellipses in the figure. Next, we perform box-focused sampling to selectively preserve object-associated blobs while effectively suppressing background noise. Finally, the sampled Gaussian blobs are fed into the detector for accurate indoor 3D object detection.
    }
    \vspace{-0.5cm}   
    \label{fig:pipeline}
\end{figure*}

\section{Related Works}
\noindent\textbf{3D Gaussian Splatting (3DGS)} is an effective explicit representation that models 3D scenes or objects using Gaussian blobs -- small, continuous Gaussian functions distributed across 3D space. Recent works~\cite{ shen2024gaussian, liu2024citygaussian, lee2024compact,zhong2025taming} have shown that 3DGS is highly suitable for dynamic scene modeling. Additionally, some studies~\cite{lin2024vastgaussian, zhang2024garfield++, xiong2024sa, wang2024pygs, liu2024efficientgs, feng2024flashgs} also demonstrate its efficiency in processing large-scale 3D scene data. 
Works~\cite{zhou2024feature, qin2024langsplat, shi2024language, zuo2024fmgs, gu2025egolifter} leverage advanced 2D foundational models, such as SAM~\cite{kirillov2023segment} and CLIP~\cite{radford2021learning}, along with feature extraction methods like DINO~\cite{zhang2022dino}, to boost perception effectiveness.
Unlike previous methods that often overlook specific challenges of indoor 3D Object Detection (3DOD), our approach uniquely introduces Boundary Guidance and Box-Focused Sampling, marking the first exploration of 3DGS as a representation for the indoor 3D object detection task.

\par\noindent\textbf{Non-View-Synthesis Representation-Based 3D Object Detection.} Traditional 3D detection tasks primarily utilize the following representations: (i) Point cloud-based methods~\cite{yang2018pixor, votenet, qi2021offboard, wang2022cagroup3d, rukhovich2022fcaf3d, cao2023coda,cao2024collaborative,yang2024imov3d} directly process 3D points captured by sensors like LiDAR or depth cameras.~Methods such as VoteNet~\cite{votenet} and CAGroup3D~\cite{wang2022cagroup3d} efficiently handle point clouds, capturing geometries while facing challenges in computational efficiency due to their irregular structure. Some works~\cite{zhou2018voxelnet, ye2020hvnet, deng2021voxel, mao2021voxel, noh2021hvpr, chen2023voxelnext, mahmoud2023dense, yan2025progressive} divide 3D space into uniform volumetric units, enabling 3D convolutional neural networks to process the data. (ii) Multi-view image-based methods~\cite{wang2022detr3d, xiong2023cape, wang2023exploring, chen2023viewpoint, feng2023aedet,  tu2023imgeonet, Shen_2024_CVPR} leverage 2D images from multiple views to reconstruct 3D structures.

\par\noindent\textbf{View-Synthesis Representation-Based Indoor 3D Object Detection.} 
NeRF~\cite{mildenhall2021nerf} have become popular for novel-view-synthesis and have been adapted for indoor 3D Object Detection (3DOD)~\cite{xu2023nerf,nerfdetpp}. These adaptations present promising solutions for detecting 3D objects using view-synthesis representations. For instance, NeRF-RPN~\cite{nerfrpn} employs voxel representations, integrating multi-scale 3D neural volumetric features to perform category-agnostic box localization~\cite{irshad2024nerfmae,nerf-fcm,yangaussian},
rather than category-specific object detection~(our task setting). NeRF-Det~\cite{xu2023nerf} incorporates multi-view geometric constraints from the NeRF component into 3D detection. 
NeRF-Det++~\cite{nerfdetpp} improves NeRF-Det for indoor multi-view 3D detection
by adding 2D semantic supervision, perspective-aware non-uniform
sampling, and ordinal-residual depth supervision.
Notably, NeRF-RPN focuses on class-agnostic box detection~\cite{nerfrpn,irshad2024nerfmae,nerf-fcm,yangaussian}, while NeRF-Det targets class-specific object detection. Our work follows the class-specific setting of NeRF-Det. However, NeRF faces a significant challenge: its implicit nature limits its representational capacity for 3D object detection. 
3DGS~\cite{kerbl3Dgaussians} has emerged as an explicit 3D representation, effectively addressing the limitation. Motivated by that, our work introduces 3DGS into indoor 3DOD for the first time, and presents novel designs to adapt 3DGS for detection, making significant differences from NeRF-based methods~\cite{nerfrpn, xu2023nerf}.

\section{Methodology}\label{sec:methods}

The pipeline of our 3DGS-DET is illustrated in~\figrefn{fig:pipeline}. Initially, we train the 3D Gaussian Splatting (3DGS) on the input scenes using the proposed Boundary Guidance, which significantly enhances the spatial distribution of Gaussian blobs, resulting in clearer differentiation between objects and the background. Subsequently, we apply the proposed Box-Focused Sampling, which effectively preserves object-related blobs while suppressing noisy background blobs. The sampled 3DGS is then fed into the detection framework for training.
In this section, we detail our method step by step. First, we introduce the preliminary concept of 3DGS in~\secrefn{sec:methods:preliminary}. As the first to introduce 3DGS in 3D object detection, we establish the basic pipeline in~\secrefn{sec:methods:basic pipeline}, utilizing 3DGS for input and output detection predictions. We then present Boundary Guidance in~\secrefn{sec:methods:boundary-guidance}. Finally, we describe the Box-Focused Sampling Strategy in~\secrefn{sec:methods:box-focused-sampling}.

\begin{figure}[t!]
     \centering
    \begin{overpic}[width=0.95\columnwidth]
    {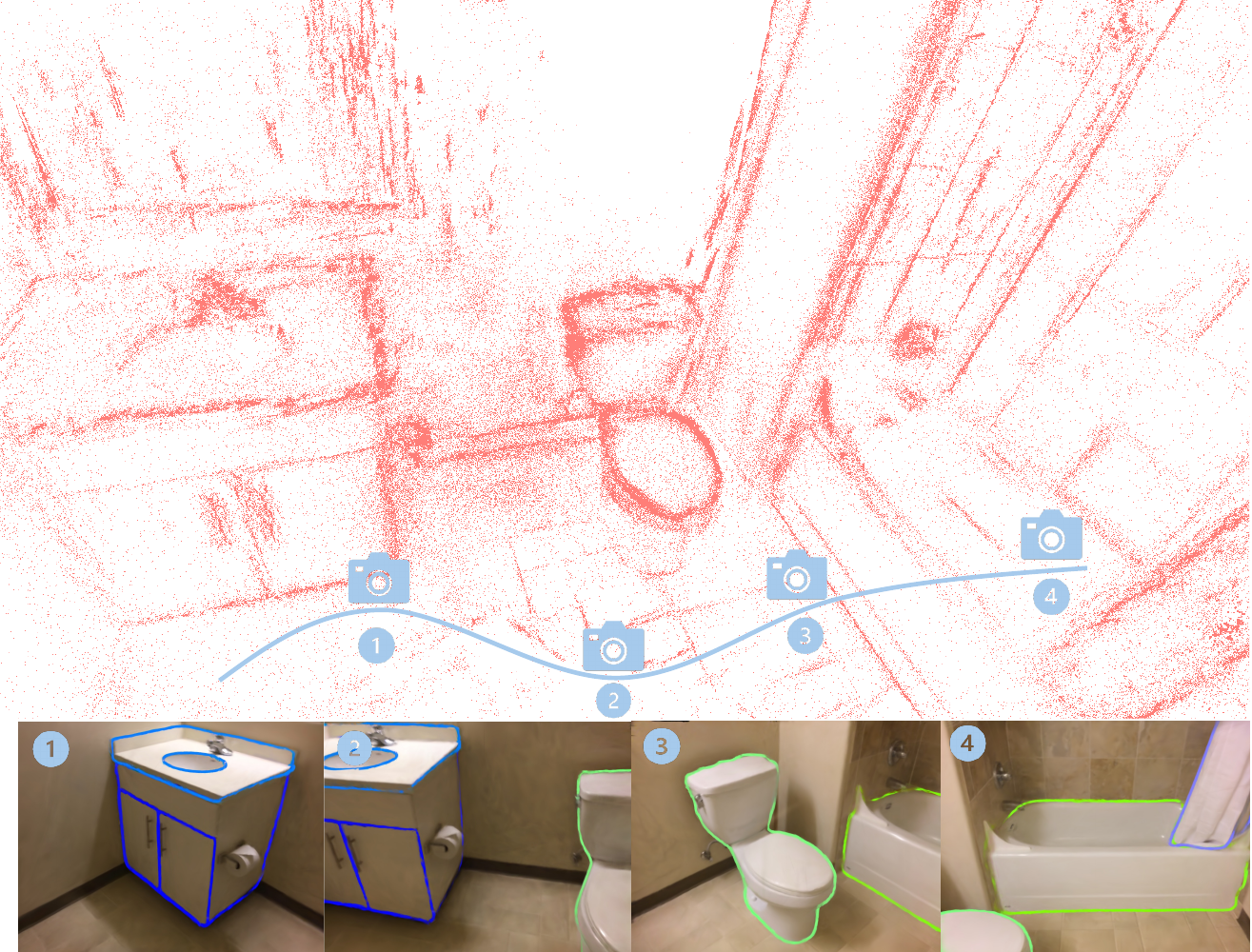}
    \end{overpic}
    \vspace{-0.2cm}  
    \caption{Rendered images from different views by 3DGS trained with Boundary Guidance. The category-specific boundaries are clearly rendered and exhibit multi-view consistency, showing that the 3D representation has successfully embedded the priors provided by Boundary Guidance.}
    \label{fig:rendered_im}
    \vspace{-0.5cm}   
\end{figure}

\subsection{Preliminary: 3D Gaussian Splatting}\label{sec:methods:preliminary}
In our proposed method, 3DGS-DET, the input scene is represented using 3DGS~\citep{kerbl3Dgaussians}, formulated as $G = \left\{ (\boldsymbol{\mu}_i, \boldsymbol{S}_i, \boldsymbol{R}_i, \boldsymbol{c}_i, \boldsymbol{\alpha}_i) \right\}_{i=1}^{N}$,
where $N$ denotes the number of Gaussian blobs. Each blob is characterized by its 3D coordinate $\boldsymbol{\mu}_i$, scaling matrix $\boldsymbol{S}_i$, rotation matrix $\boldsymbol{R}_i$, color features $\boldsymbol{c}_i$, and opacity $\boldsymbol{\alpha}_i$. These attributes define the Gaussian through a covariance matrix $\Sigma = \boldsymbol{R}\boldsymbol{S}\boldsymbol{S}^T\boldsymbol{R}^T$, centered at $\boldsymbol{\mu}$: $G(\boldsymbol{x}) = \exp^{\left(-\frac{1}{2}(\boldsymbol{x}-\boldsymbol{\mu})^T \Sigma^{-1} (\boldsymbol{x}-\boldsymbol{\mu})\right)}$.

During rendering, opacity modulates the Gaussian. By projecting the covariance onto a 2D plane~\citep{zwicker2001ewa}, we derive the projected Gaussian, and utilize volume rendering~\citep{volume_rendering} to compute the image pixel colors: $C = \sum_{k=1}^{K} \alpha_k c_k \prod_{j=1}^{k-1} (1 - \alpha_j)$,
where $K$ is the number of sampling points along the ray. $\alpha_i$ is determined by evaluating a 2D Gaussian with covariance $\Sigma$, multiplied by the learned opacity~\citep{yifan2019differentiable}. The initial 3D coordinates of each Gaussian are based on Structure from Motion (SfM) points~\citep{schonberger2016structure}. Gaussian attributes are refined to minimize the image reconstruction loss: $L_{\text{render}} = (1-\lambda) L_1(I, \hat{I}) + \lambda L_{\text{D-SSIM}}(I, \hat{I})$,
where $\hat{I}$ represents the ground truth images. Additional details can be found in~\cite{kerbl3Dgaussians}.

\subsection{Proposed Basic Pipeline of 3DGS for 3DOD}\label{sec:methods:basic pipeline}

In this section, we build our basic pipeline by directly utilizing the original 3DGS for indoor 3D Object Detection (3DOD) without any further improvement. First, we train the 3DGS representation of the input scene using posed images, denoted as $G = \left\{ (\boldsymbol{\mu}_i, \boldsymbol{S}_i, \boldsymbol{R}_i, \boldsymbol{c}_i, \boldsymbol{\alpha}_i) \right\}_{i=1}^{N}$.
Given that the number of Gaussian blobs $N$ is too large for them to be input into the detector, we perform random sampling to select a subset of Gaussian blobs, denoted as $\hat{G} = \left\{ (\boldsymbol{\mu}_i, \boldsymbol{S}_i, \boldsymbol{R}_i, \boldsymbol{c}_i, \boldsymbol{\alpha}_i) \right\}_{i=1}^{M}$, where $M < N$.
We then concatenate the attributes of the Gaussian blobs along the channel dimension as follows:
\begin{equation}
\setlength{\abovedisplayskip}{6pt}
\setlength{\belowdisplayskip}{6pt}
\hat{G}_{\text{input}} = \text{Concat}(\boldsymbol{\mu}_i, \boldsymbol{S}_i, \boldsymbol{R}_i, \boldsymbol{c}_i, \boldsymbol{\alpha}_i) \quad  \forall i \in \{1, \ldots, M\}. \label{equ:concat}
\end{equation}
This concatenated representation $\hat{G}_{\text{input}}$ is then fed into the subsequent detection tool. Note that since 3DGS is an explicit 3D representation, $\hat{G}_{\text{input}}$ can be utilized with any point-cloud-based detector by retraining the detector model on 3DGS representation. In our study, the research focus is on enhancing 3DGS for 3DOD in general, rather than designing a specific detector. Therefore, we utilize the existing work~\citep{rukhovich2022fcaf3d} as the detection tool.
The final detection predictions are obtained as follows:
\begin{equation}
\setlength{\abovedisplayskip}{6pt}
\setlength{\belowdisplayskip}{6pt}
P = \mathrm{F}(\hat{G}_{\text{input}}) = \left(\boldsymbol{p}, \boldsymbol{z}, \boldsymbol{b}\right), \label{equ:pred}
\end{equation}
where $\mathrm{F}$ denotes the detector tool and $P$ represents the predictions, including classification probabilities $\boldsymbol{p}$, centerness $\boldsymbol{z}$, and bounding box regression parameters $\boldsymbol{b}$.
The training loss~\citep{rukhovich2022fcaf3d} is defined as:
\begin{align}
\setlength{\abovedisplayskip}{6pt}
\setlength{\belowdisplayskip}{6pt}
L_{\text{det}} = &\frac{1}{N_\text{pos}} \sum_{\hat{x},\hat{y},\hat{z}} \Big(
 \mathds{1}_{\{p(\hat{x},\hat{y},\hat{z}) \neq 0\}} L_\text{reg}(\hat{\boldsymbol{b}}, \boldsymbol{b}) \notag \\
&+ \mathds{1}_{\{p(\hat{x},\hat{y},\hat{z}) \neq 0\}} L_\text{cntr}(\hat{\boldsymbol{z}}, \boldsymbol{z}) 
+  L_\text{cls}(\hat{\boldsymbol{p}}, \boldsymbol{p}) \Big),
\label{equ:loss}
\end{align}
where the number of matched positions $N_{\text{pos}}$ is given by $\sum_{\hat{x}, \hat{y}, \hat{z}} \mathds{1}_{\{p(\hat{x}, \hat{y}, \hat{z}) \neq 0\}}$. Ground truth labels are indicated with a hat symbol. The regression loss $L_{\text{reg}}$ is based on Intersection over Union (IoU), the centerness loss $L_{\text{cntr}}$ uses binary cross-entropy, and the classification loss $L_{\text{cls}}$ employs focal loss. Further details on the detection tool can be found in \cite{rukhovich2022fcaf3d}.
Building upon this basic pipeline, we develop our method, 3DGS-DET, by introducing two novel designs~(\secrefn{sec:methods:boundary-guidance} and \secrefn{sec:methods:box-focused-sampling}) to improve the 3DGS representation, as illustrated in~\figrefn{fig:pipeline}.

\begin{figure}[t!]
     \centering
    \begin{overpic}[width=0.95\columnwidth]
    {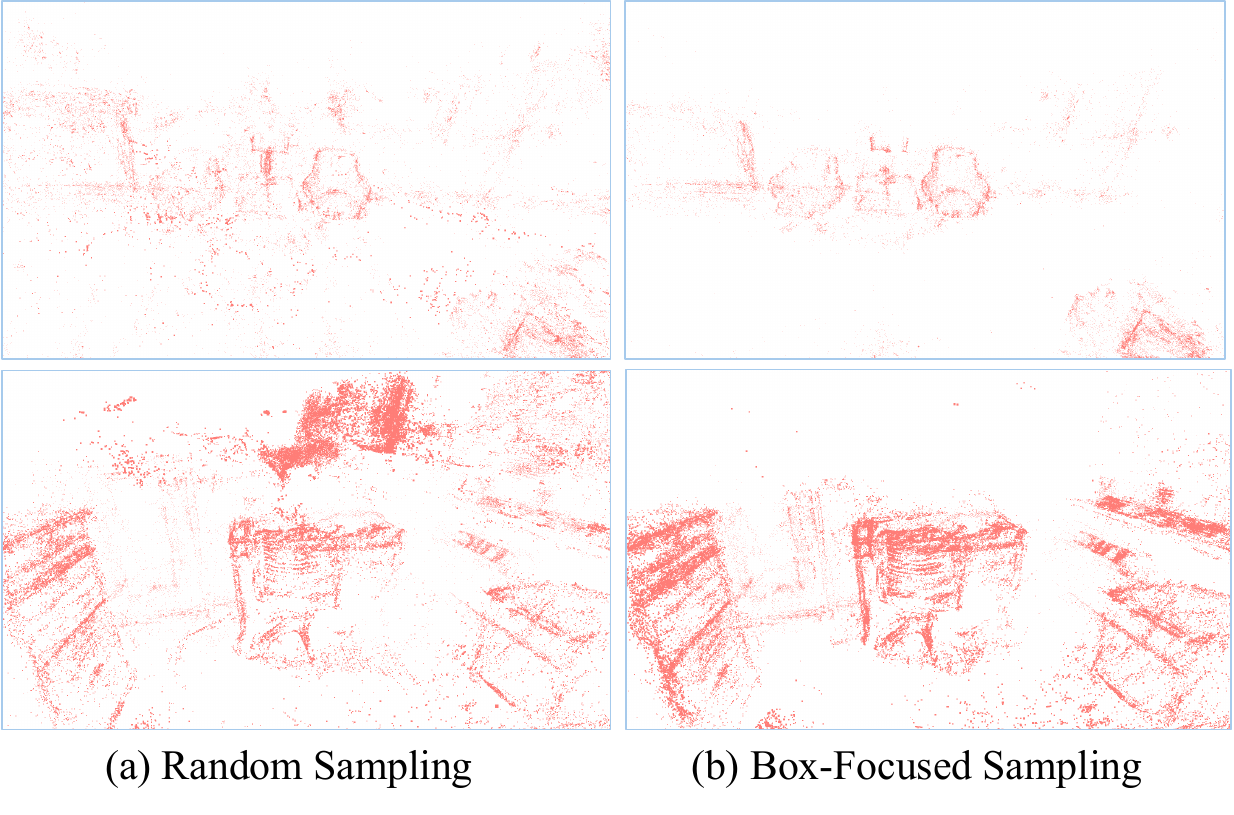}
    \end{overpic}
    \vspace{-0.3cm} 
    \caption{Qualitative Analysis of Box-Focused Sampling for 3DGS Blobs. Box-Focused Sampling significantly retains more object blobs and reduces noisy background blobs. Note that we visualize only the positions of the Gaussian blobs to highlight their spatial distribution, while omitting other attributes for clarity.}
    \label{fig:cmp_sampling}
    \vspace{-0.7cm}   
\end{figure}

\subsection{Boundary Guidance}\label{sec:methods:boundary-guidance}
Given the fact that 3DGS reconstruction
is derived from 2D images, we design the novel Boundary Guidance strategy by incorporating 2D Boundary Guidance
to achieve a more suitable 3D spatial distribution of Gaussian blobs for detection.~In this section, we present our Boundary Guidance strategy.
As illustrated in the blue block in the top row of~\figrefn{fig:pipeline}, 
to provide the guidance priors for 3DGS reconstruction, we first generate category-specific boundaries for posed images:
\begin{equation}
\setlength{\abovedisplayskip}{6pt}
\setlength{\belowdisplayskip}{6pt}
B_\text{bd} = \mathrm{H}_\text{bd}(I) = \{b_\text{bd}^c\} \quad c \in C,
\end{equation}
where $\mathrm{H}_\text{bd}$ is the boundary generator, and $b_\text{bd}^c$ represents the binary boundary map for category $c$. If $b_\text{bd}^c(x,y)=1$, the pixel at $(x, y)$ belongs to the boundary for objects of category $c$. The set $C$ includes all categories.
In practice, the operations of $\mathrm{H}_\text{bd}$ are as follows: we use Grounded SAM \citep{ren2024grounded} to generate category-specific masks. Then, the Suzuki-Abe algorithm~\citep{suzuki1985topological} is employed to extract the boundaries of these masks, along with category information. The category-specific boundaries are then overlaid on the posed images in different colors:
\setlength{\abovedisplayskip}{6pt}
\setlength{\belowdisplayskip}{6pt}   
\begin{align}
I_\text{bd}(x, y) = &I(x, y) \cdot \left(1 - \sum_{c \in C} b_\text{bd}^c(x, y)\right) \notag \\
&+ \sum_{c \in C} b_\text{bd}^c(x, y) \cdot \text{color}(c),
\label{equ:overlaid_boundary}
\end{align}
where $I_\text{bd}(x, y)$ is the pixel at position $(x, y)$ of the final image with overlaid boundaries. $I(x, y)$ is from the original image, $b_\text{bd}^c(x, y)$ is the boundary map for category $c$, and $\text{color}(c)$ is the color associated with category $c$.
These $I_\text{bd}$ images are used as ground truth to train the 3DGS representation $G_\text{bd}$
by the following loss:
\begin{equation}
\setlength{\abovedisplayskip}{6pt}
\setlength{\belowdisplayskip}{6pt}
L_{\text{render}} = (1-\lambda)L_1(I, I_\text{bd}) + \lambda L_{\text{D-SSIM}}(I, I_\text{bd}).
\label{equ:rendering_loss_2}
\end{equation}
To effectively reduce $L_{\text{render}}$ during training, it is crucial to ensure the rendering quality of boundaries and the multi-view stability of boundaries. In this way,
the Boundary Guidance lead 3DGS to incorporate boundary prior information into the 3D space.
As shown in~\figrefn{fig:teaser}, 3DGS trained with Boundary Guidance demonstrates improved spatial distribution of Gaussian blobs compared to those trained without it.
Besides,
\figrefn{fig:rendered_im}
shows the rendered images from different views by 3DGS trained
with Boundary Guidance.
The category-specific boundaries are clearly rendered with preserved multi-view consistency, evidencing that the 3D representation has successfully embedded the priors from Boundary Guidance.

\begin{table*}[t!]
\setlength{\heavyrulewidth}{1.25pt}
\centering
\resizebox{\textwidth}{!}{
\begin{tabular}{l|c|c|c|c|c|c|c|c|c|c|c|c|c|c|c|c|c|c|c}
\toprule
\textbf{Methods} & cab & bed & chair & sofa & tabl & door & wind & bkshf & pic & cntr & desk & curt & fridg & showr & toil & sink & bath & ofurn & \textbf{mAP@0.25} \\
\midrule
VoteNet & 36.3 & 87.9 & 88.7 & 89.6 & 58.8 & 47.3 & 38.1 & 44.6 & 7.8 & 56.1 & 71.7 & 47.2 & 45.4 & 57.1 & 94.9 & 54.7 & 92.1 & 37.2 & 58.7 \\
FCAF3D  & 57.2 & 87.0 & 95.0 & 92.3 & 70.3 & 61.1 & 60.2 & 64.5 & 29.9 & 64.3 & 71.5 & 60.1 & 52.4 & 83.9 & 99.9 & 84.7 & 86.6 & 65.4 & 71.5 \\
CAGroup3D & 60.4 & 93.0 & 95.3 & 92.3 & 69.9 & 67.9 & 63.6 & 67.3 & 40.7 & 77.0 & 83.9 & 69.4 & 65.7 & 73.0 & 100.0 & 79.7 & 87.0 & 66.1 & 75.12 \\
\midrule
ImGeoNet  & 40.6 & 84.1 & 74.8 & 75.6 & 59.9 & 40.4 & 24.7 & 60.1 & 4.2 & 41.2 & 70.9 & 33.7 & 54.4 & 47.5 & 95.2 & 57.5 & 81.5 & 36.1 & 54.6 \\ 
CN-RMA  & 42.3 & 80.0 & 79.4 & 83.1 & 55.2 & 44.0 & 30.6 & 53.6 & 8.8 & 65.0 & 70.0 & 44.9 & 44.0 & 55.2 & 95.4  & 68.1 & 86.1 & 49.7 & 58.6 \\ 
ImVoxelNet & 30.9 & 84.0 & 77.5 & 73.3 & 56.7 & 35.1 & 18.6 & 47.5 & 0.0 & 44.4 & 65.5 & 19.6 & 58.2 & 32.8 & 92.3 & 40.1 & 77.6 & 28.0 & 49.0 \\
\midrule
NeRF-Det  & 37.6 & 84.9 & 76.2 & 76.7 & 57.5 & 36.4 & 17.8 & 47.0 & 2.5 & 49.2 & 52.0 & 29.2 & 68.2 & 49.3 & 97.1 & 57.6 & 83.6 & 35.9 & 53.3  \\
NeRF-Det++  & 36.1 & 82.9 & 74.9 &79.1 & 57.0 & 37.3 & 24.9 & 54.6 & 2.4 & 51.7 & 72.2 & 25.5 & 58.7 & 51.5 & 92.7 & 50.8 & 82.2 & 35.1 & 53.9  \\
\rowcolor{orange!10}
3DGS-DET  & 44.1 & 82.7 & 81.7 & 79.6 & 56.0 & 35.4 & 27.6 & 45.2 & 17.3 & 61.9 & 72.8 & 40.7 & 56.6 & 71.9 & 98.5 & 72.2 & 88.3 & 46.7 & \textbf{59.9~\small{\textcolor{black}{(+6.0)}}} \\
\bottomrule
\end{tabular}}
\captionsetup{aboveskip=3pt, belowskip=0pt}
\captionof{table}{
Comparison of mAP@0.25 on the ScanNet dataset. The first two blocks include methods that utilize non-view-synthesis representations: the first block~\citep{votenet,rukhovich2022fcaf3d,wang2022cagroup3d} features approaches based on point clouds, while the second block~\citep{tu2023imgeonet,Shen_2024_CVPR,rukhovich2022imvoxelnet} focuses on multi-view image methods. The third block~\citep{xu2023nerf,nerfdetpp} encompasses methods that employ view-synthesis representations, including NeRF-based approaches and our 3DGS-based method.
Among non-view-synthesis representations, our approach clearly outperforms methods that utilize multi-view images (the second block), highlighting the superiority of our 3DGS reconstruction for detection with multi-view image inputs. In terms of view-synthesis representations, our 3DGS-DET surpasses the NeRF-based method NeRF-Det++~\cite{nerfdetpp} by 6.0 points.
}
\label{tab:main_result_0.25}
\vspace{-0.2cm} 
\end{table*}

\begin{table*}[t]
\setlength{\heavyrulewidth}{1.25pt}
\centering
\resizebox{\textwidth}{!}{ %
\begin{tabular}{l|c|c|c|c|c|c|c|c|c|c|c|c|c|c|c|c|c|c}
\toprule
\textbf{Methods} & cab & fridg & shlf & stove & bed & sink & wshr & tolt & bthtb & oven & dshw & frplce & stool & chr & tble & TV & sofa & \textbf{mAP@0.25} \\
\midrule
ImVoxelNet  
& 32.2 & 34.3 & 4.2 & 0.0 & 64.7 & 20.5 & 15.8 & 68.9 & 80.4 
& 9.9 & 4.1 & 10.2 & 0.4 & 5.2 & 11.6 & 3.1 & 35.6 & 23.6 \\

NeRF-Det  
& 36.1 & 40.7 & 4.9 & 0.0 & 69.3 & 24.4 & 17.3 & 75.1 & 84.6 
& 14.0 & 7.4 & 10.9 & 0.2 & 4.0 & 14.2 & 5.3 & 44.0 & 26.7 \\ 

NeRF-Det++  
& - & - & - & - & - & - & - & - & -
& - & - & - & - & - & - & - & - & 43.3 \\ 
\rowcolor{orange!10}
3DGS-DET~(Ours)  
& 45.2 & 84.4 & 33.3 & 41.4 & 87.3 & 75.5 & 67.6 & 87.2 & 90.8 
& 74.3 & 6.0 & 56.4 & 26.3 & 70.3 & 60.6 & 0.7 & 81.8 & \textbf{58.2 ~\small{\textcolor{black}{(+14.9)}}} \\
\bottomrule
\end{tabular}}
\captionsetup{aboveskip=3pt, belowskip=0pt}
\captionof{table}{Comparison of the `whole-scene' performance~\citep{rukhovich2022imvoxelnet,xu2023nerf} on the ARKITScenes validation set. Our 3DGS-DET significantly outperforms NeRF-Det++~\cite{nerfdetpp} by 14.9 points.~Note that we follow the setup described in the NeRF-Det supplementary materials: `In our experiments, we utilize the subset of the dataset with low-resolution images', considering it is the closest work to ours. Other methods that do not use the same setting are not listed in this table. NeRF-Det++~\cite{nerfdetpp} does not report per-category performance in ARKITScenes. We report its overall mAP@0.25 following the original paper~\cite{nerfdetpp}.}
\vspace{-0.5cm} 
\label{tab:main_result_arkit}
\end{table*}

\subsection{Box-Focused Sampling}\label{sec:methods:box-focused-sampling}
Considering that 2D images often
include numerous background pixels, leading to densely reconstructed 3DGS with
many noisy Gaussian blobs representing the background, negatively affecting detection. 
To reduce the excessive background blobs,
in this section, we propose the Box-Focused Sampling strategy in detail. As depicted in the orange block in the bottom row of~\figrefn{fig:pipeline}, 
to provide priors for the following sampling,
we utilize a 2D object detector to identify object bounding boxes:
\begin{equation}
\setlength{\abovedisplayskip}{6pt}
\setlength{\belowdisplayskip}{6pt}
B_\text{bb} = \mathrm{H_\text{bb}}(I) = \{(b_\text{bb}, p^C)\},
\end{equation}
where $\mathrm{H}_\text{bb}$ is the box detector, and we select Grounding DINO~\citep{liu2023grounding} as the detector in our experiments. Here, $b_\text{bb}$ denotes the predicted bounding box positions, and $p^C$ is the predicted probability vector for the box belonging to each category in $C$. We define $p_{\max} = \max_{c \in C} p^c$ as the highest category probability for a given bounding box, which helps to establish object probability spaces in later step. Then, 
we project the 2D boxes into 3D space:
\begin{equation}
\setlength{\abovedisplayskip}{6pt}
\setlength{\belowdisplayskip}{6pt}
F_\text{ft} = \{ K^{-1} \begin{bmatrix} x_i \\ y_i \\ z \end{bmatrix} \mid (x_i, y_i) \in b_\text{bb}, z \in \{z_{\min}, z_{\max}\} \},
\end{equation}
where $F_\text{ft}$ is the projected 3D frustum from $b_\text{bb}$, and $K^{-1}$ is the inverse camera matrix used to map 2D bounding box corners $(x_i, y_i)$ and depth values $z_{\min}$ and $z_{\max}$ into 3D space.
Next, we establish object probability spaces by $F_\text{ft}$ and $p_{\max}$. For each bounding box, the maximum probability $p_{\max}$ models the likelihood of Gaussian blobs within the corresponding frustum belonging to object blobs:
\begin{equation}
\setlength{\abovedisplayskip}{6pt}
\setlength{\belowdisplayskip}{6pt}
p_\text{obj}(g_i \mid g_i \in F_\text{ft}) = p_{\max},
\end{equation}
where $p_\text{obj}(g_i \mid g_i \in F_\text{ft})$ denotes the probability of Gaussian blob $g_i$ within frustum $F_\text{ft}$ belonging to object blobs. To integrate priors from different view frustums, we select the maximum probability as the aggregated probability:
\begin{equation}
\setlength{\abovedisplayskip}{6pt}
\setlength{\belowdisplayskip}{6pt}
p_{\text{agr}}(g_i) = \max_{v \in V} p_\text{obj}(g_i \mid g_i \in F_\text{ft}^v),
\end{equation}
where $p_{\text{agr}}(g_i)$ is the aggregated probability for Gaussian blob $g_i$, and $V$ represents the set of all views. Gaussian blobs not belonging to any frustum are assigned a small probability $p_\text{bg}$, set to 0.01 in practice.
In this way,
we obtain the object probability spaces $P_\text{obj}$, where each Gaussian blob has an associated probability of belonging to an object. We then perform probabilistic sampling based on $P_\text{obj}$ to achieve Box-Focused Sampling, resulting in the sampled Gaussian set $\hat{G}_\text{bd}^\text{bs}$ as:
\begin{equation}
\hat{G}_\text{bd}^\text{bs} = \{ g \mid g \sim P_{\text{obj}}(g) \}.
\end{equation}
In this way, it allows object blobs to be better preserved due to their higher probabilities, while most background points, having lower probabilities, are effectively reduced.
Then, based on $\hat{G}_\text{bd}^\text{bs}$, we proceed with the training of the detector, as formulated by~\equref{equ:concat}-\equref{equ:loss} as described in~\secrefn{sec:methods:basic pipeline}.
As shown in~\figrefn{fig:cmp_sampling}, 3DGS sampled via Box-Focused Sampling retains more object blobs and reduces background noise.

\section{Experiments}
\subsection{Experimental setup}
\textbf{Dataset and Metrics:} To thoroughly evaluate the performance of our proposed method in indoor 3D detection tasks, we selected two representative datasets: ScanNet~\cite{dai2017scannet} and ARKitScene~\cite{arkitscenes}. ScanNet is a large-scale indoor scene dataset containing over 1,500 real-world 3D scanned scenes, encompassing various complex indoor environments such as residential spaces, offices, and classrooms. The ARKitScene dataset is constructed from RGB-D image sequences, offering detailed geometric inform-ation and precise object annotations. For each scene, a maximum of 600 posed images are extracted. The category settings follow the standard 18 categories for ScanNet and 17 categories for ARKitScene.
We use mAP@0.25 and mAP@0.5 as the primary evaluation metrics.

\begin{figure*}[!t]
     \centering
    \begin{overpic}[width=0.95\textwidth]
    {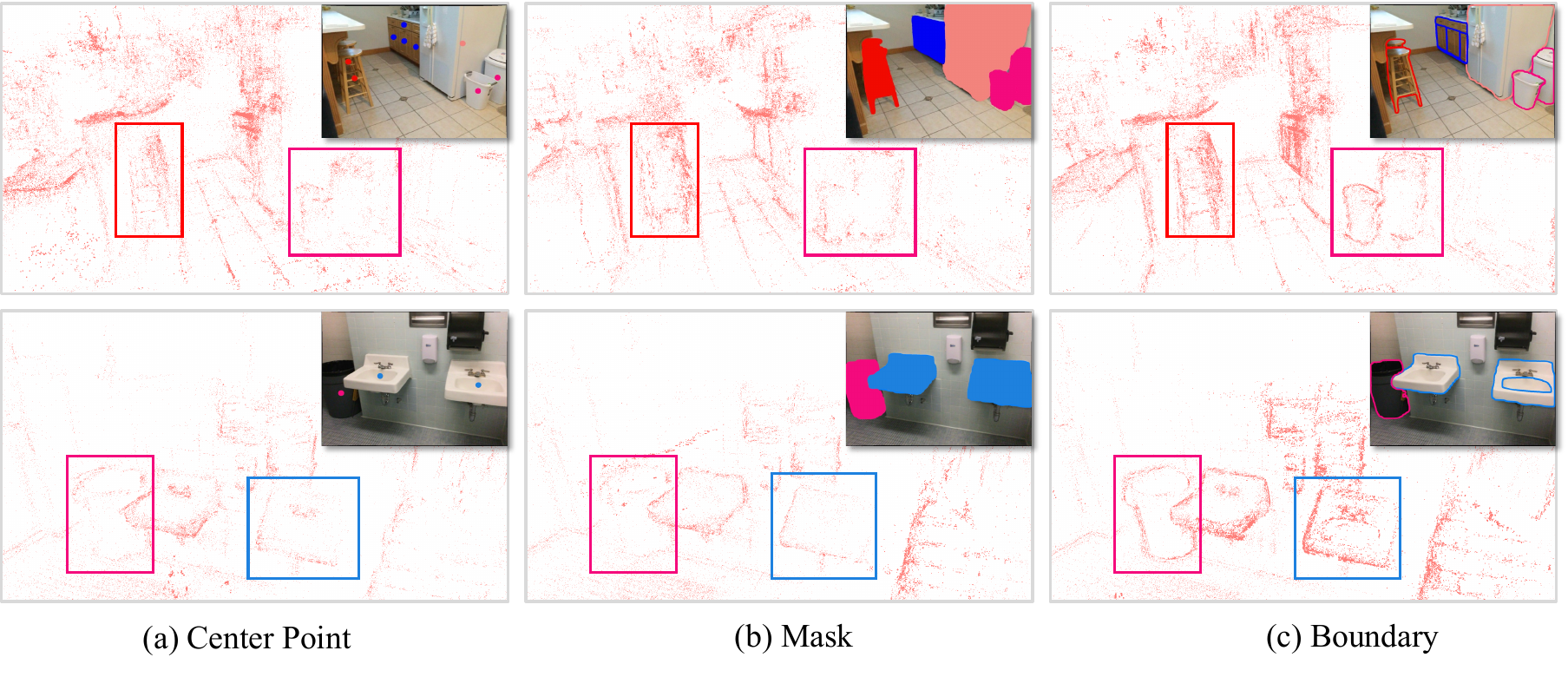}
    \end{overpic}
    \vspace{-0.3cm} 
    \caption{Analysis of guidance from different priors: (a) Center Point Guidance, (b) Mask Guidance, and (c) Boundary Guidance. In (a) and (b), the spatial distribution of Gaussian blobs for objects like the chair, trash bin and sink is incomplete and ambiguous. Gaussian blobs trained with Boundary Guidance exhibit a clearer spatial distribution. For detailed analysis, please refer to~\secrefn{sec:diff_priors}.
    }
    \vspace{-0.1cm} 
    
    \label{fig:abl_point_mask}
\end{figure*}

\begin{table*}[!t]
\setlength{\heavyrulewidth}{1.25pt}
    \centering
    \begin{minipage}{0.275\textwidth} 
        \centering
        \resizebox{\linewidth}{!}{
            \begin{tabular}{l|c|c}
                \toprule
                \textbf{Methods} & \textbf{mAP@0.25} & \textbf{mAP@0.5} \\
                \midrule
                BP & 54.3 & 34.1 \\
                BP+BG & 56.7 & 36.9 \\
                \rowcolor{orange!10} 
                BP+BG+BS  & \textbf{59.9}  & \textbf{37.8} \\
                \bottomrule
            \end{tabular}
        }
        \vspace{-0.2cm} 
        \caption{Effect of our designs.}
        \vspace{-0.5cm} 
        \label{tab:effect_design}
    \end{minipage}
    \hfill
    \begin{minipage}{0.315\textwidth}
        \centering
        \resizebox{\linewidth}{!}{
            \begin{tabular}{l|c|c}
                \toprule
                \textbf{Different Priors} & \textbf{mAP@0.25} & \textbf{mAP@0.5} \\
                \midrule
                2D Center Point & 54.4 & 33.9 \\
                2D Mask & 54.9 & 34.2 \\
                \rowcolor{orange!10} 
                2D Boundary (ours) & \textbf{56.7}  & \textbf{36.9} \\
                \bottomrule
            \end{tabular}
        }
        \vspace{-0.2cm} 
        \caption{Guidance from different priors.}
        \vspace{-0.5cm} 
        \label{tab:prior_comparison}
    \end{minipage}
    \hfill
    \begin{minipage}{0.38\textwidth}
        \centering
        \resizebox{\linewidth}{!}{
            \begin{tabular}{l|c|c}
                \toprule
                \textbf{Sampling Methods} & \textbf{mAP@0.25} & \textbf{mAP@0.5} \\
                \midrule
                Random Sampling & 56.7 & 36.9 \\
                Farthest Point Sampling  & 57.4 & 37.6 \\ 
                \rowcolor{orange!10} 
                Box-focused Sampling (ours) & \textbf{59.9} & \textbf{37.8}  \\
                \bottomrule
            \end{tabular}
        }
        \vspace{-0.2cm} 
        \caption{Different sampling methods.}
        \vspace{-0.5cm} 
        \label{tab:sample}
    \end{minipage}
\end{table*} 

\subsection{Main Results}
\textbf{Quantitative Results}.
For the \textbf{ScanNet} dataset, we present the mAP@0.25 and mAP@0.5 performances of various methods in \tabref{tab:main_result_0.25} of the main paper and Tab. 3 of the supplementary, respectively. 
In both~\tabref{tab:main_result_0.25} and Tab. 3 of the supplementary, the methods listed in the first two blocks
 are non-view-synthesis representation-based 3D detection methods:
the first block~\citep{votenet,rukhovich2022fcaf3d,wang2022cagroup3d} lists approaches based on point clouds, while the second block
focuses on multi-view image methods~\citep{tu2023imgeonet,Shen_2024_CVPR,rukhovich2022imvoxelnet}. 
The third block consists of view-synthesis representation-based 3DOD methods, including NeRF-Det~\citep{xu2023nerf}, NeRF-Det++~\citep{nerfdetpp} and our proposed 3DGS-DET. NeRF-Det and NeRF-Det++ are the closest work to ours, leveraging Neural Radiance Fields (NeRF).
`3DGS-DET' is our full method, utilizing both Boundary Guidance and Box-Focused Sampling as described in \secrefn{sec:methods:box-focused-sampling}.
As illustrated in \tabref{tab:main_result_0.25} and Tab. 3 of the supplementary, our method significantly outperforms NeRF-Det++ by \textbf{+6.0} on mAP@0.25 and \textbf{+7.8} on mAP@0.5, showcasing the superiority of our approach.
Besides, our method also clearly surpasses approaches utilizing multi-view images as input, demonstrating the superiority of our 3DGS reconstruction for detection.

Regarding the \textbf{ARKitScene} dataset, considering NeRF-Det is the closest work to ours, we follow the same setup described in the NeRF-Det~\citep{xu2023nerf} supplementary materials: `In our experiments, we utilize the subset of the dataset with low-resolution images.' Similarly, we adopt the same subset of the ARKitScenes dataset. Other methods that report performance on ARKitScene use the full dataset, so our 3DGS-DET is only compared with ImVoxelNet, NeRF-Det and NeRF-Det++ under the same conditions. The results in \tabref{tab:main_result_arkit} demonstrate that 3DGS-DET performs better across most categories, achieving an mAP@0.25 of 58.2, which significantly outperforms NeRF-Det++ by \textbf{+14.9}, highlighting the superiority of our method.

\par\noindent\textbf{Qualitative results}.
We provide a qualitative comparison with NeRF-Det in Fig. 1 and Fig. 2 of the supplementary. As shown, our method detects more objects in the scene with greater positional accuracy compared to NeRF-Det~\citep{xu2023nerf}, demonstrating the superiority of our method.

\subsection{Analysis on the Proposed Designs}   \label{sec:effect_designs}

In this section, we demonstrate the effectiveness of our contributions by first presenting the performance of our proposed basic 3DGS detection pipeline and then incrementally incorporating our additional designs to analyze the resulting performance improvements.
We further delve into the method and provide additional analyses to
for a more comprehensive understanding of our method.

\par\noindent\textbf{Our Proposed Basic 3DGS Detection Pipeline}. 
As shown in~\tabref{tab:effect_design}, `BP'
represents our basic pipeline,
i.e., our proposed detection pipeline utilizing 3DGS, as described in~\secrefn{sec:methods:basic pipeline}. Benefiting from the advantages of 3DGS as an explicit scene representation, our basic pipeline surpasses NeRF-Det++ (54.3 vs. 53.9), underscoring the significance of introducing 3DGS into indoor 3DOD for the first time.

\par\noindent\textbf{Effectiveness of Boundary Guidance}. In~\tabref{tab:effect_design}, `BP+BG' incorporates the proposed Boundary Guidance as detailed in~\secrefn{sec:methods:boundary-guidance}. Introducing Boundary Guidance into the basic pipeline results in a significant improvement of 2.4 points (56.7 vs. 54.3), demonstrating the effectiveness of the proposed Boundary Guidance.
To further explore the impact of Boundary Guidance on 3DGS representations, 
we present a visual comparison of the spatial distribution of trained Gaussian blobs in Fig. 3 in the supplementary.
As we can see, Gaussian blobs trained with Boundary Guidance demonstrate clearer spatial distribution and more distinct differentiation between objects and the background.
We also present rendered images from different views by 3DGS trained with Boundary Guidance in Fig. 4 and Fig. 5 in the supplementary.
As can be observed, the category-specific boundaries are clearly rendered and show multi-view stability, indicating that the 3D representation has effectively embedded the priors from Boundary Guidance. All these results clearly verify the effectiveness of the proposed Boundary Guidance for 3DGS-Det.

\par\noindent\textbf{Effectiveness of Box-Focused Sampling}. Furthermore, we introduce Box-Focused Sampling detailed in~\secrefn{sec:methods:box-focused-sampling}, represented by `BP+BG+BS' in~\tabref{tab:effect_design}. This addition leads to a further performance boost of 3.2 points (59.9 vs. 56.7), proving the effectiveness of Box-Focused Sampling.
The visual comparison of sampled Gaussian blobs is shown in Fig. 6 in the supplementary.
We can observe that the proposed Box-Focused Sampling significantly
retains more object blobs and suppresses noisy background blobs.

\par\noindent\textbf{Guidance from Different Priors}. \label{sec:diff_priors}
In this section, we analyze the impact of guidance from various priors. As described in~\secrefn{sec:methods:boundary-guidance}, we utilize the object's boundary as the guidance prior. Here, we perform an ablation study considering the object's center point and mask as alternative priors. To obtain the center point, we detect the object's bounding box using GroundingDINO~\citep{liu2023grounding} and compute its center coordinates. The mask is generated with GroundedSAM~\citep{ren2024grounded}. Note that all priors are category-specific, with each class associated with a fixed color. These priors are overlaid on the posed images, as shown in~\figrefn{fig:abl_point_mask}, and then used to train the 3DGS for detection. \tabref{tab:prior_comparison} presents the detection performance for 3DGS trained with the different priors.
As reported in~\tabref{tab:prior_comparison}, the 3DGS-DET method using boundary guidance achieved 56.7\% in mAP@0.25 and 36.9\% in mAP@0.5, demonstrating significant superiority over the alternative priors.

\begin{table}[t!]
\centering

\begingroup
{\footnotesize\relsize{+0.5}
\setlength{\tabcolsep}{3pt}        %
\renewcommand{\arraystretch}{0.98} %

\begin{tabularx}{\columnwidth}{l|*{2}{>{\centering\arraybackslash}X}}
\toprule
\centering \textbf{Sampling Methods} & \textbf{mAP@0.25} & \textbf{mAP@0.5} \\
\midrule
\centering Box-Focused Sampling & 59.9 & 37.8 \\
\rowcolor{orange!10}
\centering Box-Focused Sampling-GT-Box & \textbf{76.5} & \textbf{68.3} \\
\bottomrule
\end{tabularx}}
\endgroup

\begingroup
\captionsetup{aboveskip=2pt,belowskip=0pt}
\captionof{table}{%
Upper bound of Box-Focused Sampling with GT boxes.}
\label{tab:upper}
\vspace{-0.4cm}
\endgroup
\end{table}

\begin{table}[t!]
\centering

\begingroup
{\footnotesize\relsize{+0.5}
\setlength{\tabcolsep}{3pt}        %
\renewcommand{\arraystretch}{0.98} %

\begin{tabularx}{\columnwidth}{l|*{2}{>{\centering\arraybackslash}X}}
\toprule
\centering \textbf{Sampling Methods} & \textbf{Average Noisy Points} \\
\midrule
\centering Farthest Point Sampling (FPS) & 80025 \\
\rowcolor{orange!10}
\centering Box-Focused Sampling (Ours) & {\textbf{59476~{\textcolor{black}{(-25.7\%)}}}} \\
\bottomrule
\end{tabularx}}
\endgroup

\begingroup
\captionsetup{aboveskip=2pt,belowskip=0pt}
\captionof{table}{%
Average noisy points: FPS vs. Box-Focused Sampling.}
\label{tab:noise_cmp}
\vspace{-0.5cm}
\endgroup
\end{table}

Let's explore the visualizations for further insights. In (a) and (c) of~\figrefn{fig:abl_point_mask}, we observe that the spatial distribution of Gaussian blobs with Point Guidance is less distinct compared to Boundary Guidance. This is because the center point provides only positional guidance, lacking richer information like shape or size, making it less effective compared to the boundary prior.
For the mask prior, as shown in (b) and (c) of~\figrefn{fig:abl_point_mask}, the Gaussian blobs' spatial distribution with Mask Guidance is more ambiguous than with the Boundary Guidance. Although the mask highlights shape and size information, it hides the object's surface, reducing texture and geometric information, thus being less effective than the boundary prior.
Overall, Boundary Guidance offers positional cues and richer information such as shape and size while preserving texture and geometric details on the object's surface, leading to the best performance.

\par\noindent\textbf{Different Sampling Methods}.
In this section, we compare two additional sampling methods with our Box-Focused Sampling: 1) Random Sampling and 2) Farthest Point Sampling~\citep{qi2017pointnetplusplus}. The latter iteratively selects points farthest from those already chosen, ensuring even distribution for better scene coverage, focusing on global distribution rather than specific geometric features of objects.
The results in~\tabref{tab:sample} demonstrate that our Box-Focused Sampling achieves the highest performance, with mAP@0.25 and mAP@0.5 reaching 59.9\% and 37.8\%, respectively. This is because 3DGS often contain excessive background blobs. Our Box-Focused Sampling is specifically designed to preserve more object-related blobs while suppressing noisy background blobs. In contrast, other sampling methods primarily focus on global scenes without differentiation between objects and background blobs.

\par\noindent\textbf{The Upper Bound of Box-Focused Sampling}.
In this section, we conduct an ablation study on the upper bound of Box-Focused Sampling by utilizing ground-truth~(GT) 3D bounding boxes to sample 3DGS blobs. As shown in~\tabref{tab:upper}, despite the significant improvements achieved with Box-Focused Sampling~(3.2 points in~\tabref{tab:effect_design}), using GT 3D boxes brings further enhancements, highlighting its potential as a promising research direction.

\begin{table}[t!]
\centering

\begingroup
{\footnotesize\relsize{+0.5}
\setlength{\tabcolsep}{3pt}        %
\renewcommand{\arraystretch}{0.98} %

\begin{tabularx}{\columnwidth}{l|*{2}{>{\centering\arraybackslash}X}}
\toprule
\textbf{Class-agnostic Setting} & \textbf{mAP@0.25} & \textbf{mAP@0.5} \\
\midrule
NeRF-RPN~\cite{nerfrpn}  & 55.5 & 18.4 \\
NeRF-MAE~\cite{irshad2024nerfmae} & 57.1 & 17.0 \\
NeRF-FCM~\cite{nerf-fcm} & 58.8 & 23.4 \\
Gaussian-Det~\cite{yangaussian} & 71.7 & 24.5  \\
\rowcolor{orange!10}
3DGS-DET (ours) & \textbf{75.6 ~{\textcolor{black}{(+3.9)}}} & \textbf{52.3 ~{\textcolor{black}{(+27.8)}}} \\
\bottomrule
\end{tabularx}}
\endgroup

\begingroup
\captionsetup{aboveskip=2pt,belowskip=0pt}
\captionof{table}{%
Performance on the class-agnostic setting~\cite{nerfrpn,irshad2024nerfmae,nerf-fcm,yangaussian}, 
which targets class-agnostic box detection, i.e., localizing objects 
without predicting their semantic categories.}
\label{tab:rpn}
\vspace{-0.5cm}
\endgroup
\end{table}

\par\noindent\textbf{Noise Reduction via Box-Focused Sampling}.
We compare the average number of noisy points in test scenes on ScanNet when using Farthest Point Sampling (FPS) and our Box-Focused Sampling. As shown in~\tabref{tab:noise_cmp}, our Box-Focused Sampling
significantly reduces noisy background blobs by \textbf{25.7\%} compared with the FPS sampling,
demonstrating its effectiveness in noise reduction.

\subsection{Performance on NeRF-RPN Setting}

In this section, we adapt our 3DGS-DET to the NeRF-RPN setting~\citep{nerfrpn,irshad2024nerfmae}, which targets class-agnostic box detection. To achieve this, we labeled all the ground-truth boxes with a single `object' category and trained 3DGS-DET accordingly. 
Additionally, NeRF-RPN uses a different train/validation split compared to the official ScanNet dataset, with its validation set overlapping the official ScanNet training set. To address this, we excluded the overlapping parts between the NeRF-RPN validation set and the ScanNet official training set from our training data. We then used the remaining scenes for training, and evaluated on the same validation set provided by NeRF-RPN.
As shown in~\tabref{tab:rpn}, 3DGS-DET achieved an mAP@0.25 of 75.6\% and an mAP@0.5 of 52.3\%, significantly outperforming Gaussian-Det~\cite{yangaussian}'s 71.7\% and 24.5\%. This demonstrates the significant superiority of our method in the class-agnostic setting.

\subsection{More Ablation Studies}
To comprehensively analyze our method, in Sec. 2 of the supplementary material, we further study `The Effect of Boundary Guidance on the Spatial Distribution of 3DGS Blobs',  `Comparison with SfM-Based 3D Detection Baseline', `Application to Feed-Forward 3DGS', `Efficiency Analysis' , `Failure Cases' and provide extensive qualitative analysis.

\section{Conclusion}\label{sec:conclusion}

In this work, we introduce 3DGS into indoor 3D Object Detection~(3DOD) for the first time and propose 3DGS-DET, a novel approach that leverages Boundary Guidance and Box-Focused Sampling to enhance 3DGS for indoor 3DOD. Our method addresses 3DGS challenges by improving spatial distribution and reducing background noise. With 2D Boundary Guidance, we achieve clearer object-background differentiation, while Box-Focused Sampling retains more object points and suppresses noise. 3DGS-DET significantly outperforms state-of-the-art methods.

{
    \small
    \bibliographystyle{ieeenat_fullname}
    \bibliography{main}
}

\clearpage
\setcounter{section}{0} %
\setcounter{subsection}{0}
\setcounter{figure}{0}
\setcounter{table}{0}
\renewcommand\thesection{\arabic{section}} %

\begin{center}
\LARGE\bfseries Supplementary Material
\end{center}

\section{Implementation Details} 
To train 3DGS, we follow~\cite{kerbl3Dgaussians} to initialize the 3D coordinates of Gaussian blobs using Structure-from-Motion (SfM) points. The training hyperparameters are the same as those in~\cite{kerbl3Dgaussians}. We employ 
pretrained GroundedSAM~\citep{ren2024grounded} and the Suzuki-Abe algorithm~\citep{suzuki1985topological} as the boundary detector in Boundary Guidance. The pretrained GroundingDINO~\citep{liu2023grounding} is used as the box detector in the Box-Focused Sampling strategy.
Note that GroundingDINO is integrated as part of GroundedSAM. Therefore, in practice, only GroundedSAM needs to be executed to obtain both predictions.
For the detection tool, we utilize FCAF3D~\citep{rukhovich2022fcaf3d} implemented in MMDetection3D~\citep{mmdet3d2020}. The training hyperparameters are the same as those in FCAF3D. In our ablation study, to ensure a fair comparison, all model versions are trained with the same hyperparameters, such as the same number of epochs, specifically 12 epochs. All the ablation experiments~(Sec.~4.3 in the main paper) are conducted on ScanNet.

\section{Additional Ablation Studies} 

\subsection{The Effect of Boundary Guidance on the Spatial Distribution of 3DGS Blobs} %
\textbf{Quantitative Study.} To investigate the effect of Boundary Guidance on the spatial distribution of Gaussian blobs, we conducted an ablation study utilizing the first 100 scenes of ScanNet. Specifically, we evaluated the spatial distribution of Gaussian blobs by calculating the \textit{Root Mean Square Error (RMSE) metric}~\cite{zhang20233d} between the Gaussian blob positions and the ground-truth point clouds. Without Boundary Guidance, the average RMSE was 56.08\%, whereas incorporating Boundary Guidance reduced the RMSE to 53.99\%, achieving a significant improvement of \textbf{2.09 points}. This result demonstrates that Boundary Guidance effectively improves
the spatial distribution of Gaussian blobs.

\noindent\textbf{Qualitative Study.}
In addition to the quantitative evaluation, we provide a qualitative analysis of the spatial distribution of Gaussian blobs in~\figrefn{fig:guidance_supp}. Gaussian blobs trained with Boundary Guidance demonstrate clearer spatial distributions and more distinct
differentiation between objects and the background. These qualitative results visually confirm the quantitative findings above, illustrating the effectiveness of Boundary Guidance in improving the spatial distribution of Gaussian blobs.

\subsection{Rendered Images from 3DGS Trained with Boundary Guidance}
We present rendered images from different views generated by 3DGS trained with Boundary Guidance in~\figrefn{fig:render 1} and~\figrefn{fig:render 2}. As shown, the category-specific boundaries are clearly rendered and exhibit multi-view consistency, demonstrating that the 3D representation effectively incorporates the priors from Boundary Guidance. 

\subsection{Qualitative Analysis of Box-Focused Sampling for 3DGS Blobs}
\figrefn{fig:sample_supp} visually demonstrates the effect of Box-Focused Sampling. As shown, Box-Focused Sampling enhances the preservation of object blobs while effectively reducing noisy background blobs.

\subsection{Comparison with SfM-based 3D Detection Baseline}
We implement the baseline combining SfM~\cite{schonberger2016structure,schonberger2016pixelwise} points with the FCAF3D~\cite{rukhovich2022fcaf3d}.
As shown in~\tabref{tab:sfm}, our
method significantly outperforms
the SfM-Detector by 11.1 points regarding mAP@0.25 and 8.1 points regarding mAP@0.5.
The performance gap primarily stems from SfM reconstructions' sparse point distributions. In contrast, our 3DGS with Boundary Guidance and Box-Focused Sampling promote superior spatial distribution of dense Gaussian blobs while suppressing background noise, ultimately yielding significant performance gains.

\begin{table}[t!]
\centering
\resizebox{0.8\columnwidth}{!}{
\begin{tabular}{l|c|c}
\toprule
Method & \textbf{mAP@0.25} & \textbf{mAP@0.5} \\
\midrule
NeRF-Det~\cite{xu2023nerf} & 53.3  & 29.7 \\
NeRF-Det++~\cite{nerfdetpp} & 53.9  & 30.0 \\
\rowcolor{orange!10} 3DGS-DET~(FF)       & \textbf{58.5~\small{\textcolor{black}{(+4.6)}}} & \textbf{31.8~\small{\textcolor{black}{(+1.8)}}} \\ 
\bottomrule
\end{tabular}
}
\caption{When applied to the feed-forward 3DGS, the performance of our method (`3DGS-DET~(FF)' still remains significantly ahead.}
\label{tab:ff_comparison}
\end{table}

\subsection{Application to Feed-forward 3DGS}
Vanilla 3DGS~\citep{kerbl3Dgaussians} optimizes each training scene individually, whereas the detector is trained on the entire training dataset. This fundamental difference in the optimization process prevents end-to-end training.
However, to achieve end-to-end training, our method is flexible to be applied not only to Vanilla 3DGS but also to more recent feed-forward 3DGS methods~\cite{chen2024mvsplat}, which are trained on entire datasets.
We specifically applied our designs to the feed-forward 3DGS in an end-to-end integration manner and still achieved much better results than NeRF-Det++~\cite{nerfdetpp}, as shown in \tabref{tab:ff_comparison}.

\subsection{Efficiency analysis} When applied to the feed-forward 3DGS, our method (`3DGS-DET~(FF)' in \tabref{tab:ff_comparison}) completes the entire training phase in 11 hours and the testing phase in 75 seconds on ScanNet, delivering significant performance improvements over NeRF-Det++ with an acceptable computational cost. All timing statistics were measured on two A800 GPUs.

\begin{table}[t!]
\centering
    \centering
    \begin{tabular}{l|c|c}
    \toprule
    $P_\text{obj}$ & \textbf{mAP@0.25} & \textbf{mAP@0.5} \\
    \midrule
    SfM-Detector & 48.8 & 29.7 \\
     \rowcolor{orange!10} Ours &  \textbf{59.9~\small{\textcolor{black}{(+11.1)}}} &  \textbf{37.8~\small{\textcolor{black}{(+8.1)}}} \\
    \bottomrule
    \end{tabular}
    \captionsetup{aboveskip=2pt}\captionsetup{belowskip=0pt}
    \caption{Comparison with SfM-based 3D Detection Baseline~\cite{schonberger2016structure,schonberger2016pixelwise,rukhovich2022fcaf3d}.}
    \label{tab:sfm}
\end{table}

\begin{table*}[t!]
\centering
\resizebox{\textwidth}{!}{
\begin{tabular}{l|c|c|c|c|c|c|c|c|c|c}
\toprule
\textbf{Methods} & cab & bed & chair & sofa & tabl & door & wind & bkshf & pic & cntr \\
\midrule
VoteNet \citep{votenet} & 8.1 & 76.1 & 67.2 & 68.8 & 42.4 & 15.3 & 6.4 & 28.0 & 1.3 & 9.5 \\
FCAF3D \citep{rukhovich2022fcaf3d} & 35.8 & 81.5 & 89.8 & 85.0 & 62.0 & 44.1 & 30.7 & 58.4 & 17.9 & 31.3 \\
CAGroup3D \citep{wang2022cagroup3d} & 41.4 & 82.8 & 90.8  & 85.6 & 64.9 & 54.3 & 37.3 & 64.1 & 31.4 & 41.1 \\
\midrule
ImGeoNet \citep{tu2023imgeonet} &15.8 & 74.8 & 46.5 & 45.7 & 39.9 & 8.0 & 2.9 & 32.9 & 0.3 & 7.9 \\
CN-RMA \citep{Shen_2024_CVPR} & 21.3 & 69.2 & 52.4 & 63.5 & 42.9  & 11.1 & 6.5 & 40.0 & 1.2 & 24.9 \\
ImVoxelNet~\citep{rukhovich2022imvoxelnet} & 8.9 & 67.1 & 35.0 & 33.1 & 30.5 & 4.9 & 1.3 & 7.0 & 0.1 & 0.9 \\

\midrule
NeRF-Det~\citep{xu2023nerf} & 12.0  & 68.4 & 47.8 & 58.3 & 42.8 & 7.1 & 3.0  & 31.3 & 1.6 & 11.6 \\
NeRF-Det++~\citep{nerfdetpp}  & - & - & - & - & - & - & - & - & \multicolumn{2}{c}{- } \\
\rowcolor{orange!10} 
3DGS-DET~(Our basic pipeline) & 18.5 & 73.5 & 44.6 & 61.9 & 42.2 & 9.3 & 5.6 & 28.7 & 2.3 & 2.0 \\
\rowcolor{orange!10} 
3DGS-DET~(Our basic pipeline+BG) & 16.1 & 77.0 & 51.6 & 62.4 & 44.7 & 11.7 & 11.3 & 24.4 & 1.7 & 19.0 \\
\rowcolor{orange!10} 
3DGS-DET~(Our basic pipeline+BG+BS) & 19.2 & 73.8 & 52.7 & 65.2 & 46.2 & 9.6 & 8.2 & 31.8 & 4.2 & 20.9 \\
\toprule 
\textbf{Methods} & desk & curt & fridg & showr & toil & sink & bath & ofurn & \multicolumn{2}{c}{\textbf{mAP@0.5}} \\
\midrule
VoteNet \citep{votenet} & 37.5 & 11.6 & 27.8 & 10.0 & 86.5 & 16.8 & 78.9 & 11.7 & \multicolumn{2}{c}{ 33.5 } \\
FCAF3D \citep{rukhovich2022fcaf3d} & 53.4 & 44.2 & 46.8 & 64.2 & 91.6 & 52.6 & 84.5 & 57.1 & \multicolumn{2}{c}{ 57.3 } \\
CAGroup3D \citep{wang2022cagroup3d} & 63.6 & 44.4 & 57.0 & 49.3 & 98.2 & 55.4 & 82.4 & 58.8 & \multicolumn{2}{c}{ 61.3 } \\
\midrule
ImGeoNet \citep{tu2023imgeonet} & 43.9 & 4.3 & 24.0 & 2.0 & 68.8 & 24.5 & 61.7 & 17.4 & \multicolumn{2}{c}{28.9} \\ %
CN-RMA \citep{Shen_2024_CVPR} & 51.4 & 19.6 & 33.0 & 6.6 & 73.3  & 36.1 & 76.4 & 31.5 & \multicolumn{2}{c}{36.8} \\
ImVoxelNet~\citep{rukhovich2022imvoxelnet} & 35.5 & 0.6 & 22.1 & 4.5 & 67.7 & 18.9 & 60.2 & 10.1 & \multicolumn{2}{c}{22.7} \\

\midrule
NeRF-Det~\citep{xu2023nerf}  & 46.0 & 5.8 & 26.0 & 1.6 & 69.0 & 25.5 & 55.8 & 21.1 & \multicolumn{2}{c}{29.7 } \\
NeRF-Det++~\citep{nerfdetpp}  & - & - & - & - & - & - & - & - & \multicolumn{2}{c}{30.0 } \\
\rowcolor{orange!10}
3DGS-DET~(Our basic pipeline)& 53.5 & 18.1 & 30.7 & 3.4 & 77.0 & 29.0 & 68.3 & 24.2 & \multicolumn{2}{c}{34.1} \\
\rowcolor{orange!10}
3DGS-DET~(Our basic pipeline+BG) & 47.4 & 27.2 & 30.4 & 8.3 & 87.0 & 36.3 & 78.3 & 28.8 & \multicolumn{2}{c}{36.9} \\
\rowcolor{orange!10} 
3DGS-DET~(Our basic pipeline+BG+BS) & 52.4 & 22.2 & 36.9 & 15.7 & 82.6 & 35.1 & 74.0 & 28.9 & \multicolumn{2}{c}{\textbf{37.8~\small{\textcolor{black}{(+7.8)}}}} \\
\bottomrule
\end{tabular}}
\captionsetup{aboveskip=2pt}\captionsetup{belowskip=0pt}\captionof{table}{Comparison of mAP@0.5 on ScanNet. The first two blocks include methods that utilize non-view-synthesis representations: the first block consists of methods based on point clouds~\cite{votenet,rukhovich2022fcaf3d,wang2022cagroup3d}, while the second block focuses on multi-view image methods~\cite{tu2023imgeonet,Shen_2024_CVPR,rukhovich2022imvoxelnet}. The third block includes methods that employ view-synthesis representations, including NeRF-based approaches and our 3DGS-based method.
Among non-view-synthesis representations, our approach clearly outperforms methods that utilize multi-view images (the second block), demonstrating the superiority of our 3DGS reconstruction for detection with multi-view image inputs. In terms of view-synthesis representations, our 3DGS-DET significantly surpasses the NeRF-based method NeRF-Det++~\citep{nerfdetpp} by 7.8 points. NeRF-Det++~\cite{nerfdetpp} does not report per-category performance of mAP@0.5. We report its overall mAP@0.5 following the original paper~\cite{nerfdetpp}.}
\label{tab:main_result_0.5}
\end{table*}

\section{Additional Main Results} 
\subsection{Additional Quantitative Results}
In \tabref{tab:main_result_0.5},
we compare the mAP@0.5 performance of various methods on the ScanNet dataset. 
The methods are grouped into three blocks based on their underlying representations. 
The first two blocks include methods that utilize non-view-synthesis representations: the first block consists of methods based on point clouds~\cite{votenet,rukhovich2022fcaf3d,wang2022cagroup3d}, while the second block focuses on multi-view image methods~\cite{tu2023imgeonet,Shen_2024_CVPR,rukhovich2022imvoxelnet}. The third block includes methods that employ view-synthesis representations, including NeRF-based methods~\cite{xu2023nerf,nerfdetpp} and our 3DGS-based method.
Among non-view-synthesis representations, our method achieves 
clearly superior performance compared to methods utilizing multi-view images (the second block), demonstrating the effectiveness of our 3DGS reconstruction for detection tasks with multi-view image inputs. As for the view-synthesis representations, our 3DGS-DET outperforms the NeRF-based method NeRF-Det++~\cite{nerfdetpp} by a significant margin of \textbf{\textcolor{black}{7.8} points}, further highlighting the advantages of our 3DGS-DET.

\subsection{Additional Qualitative results}
\begin{figure*}[ht]
     \centering
    \begin{overpic}[width=0.9\textwidth]
    {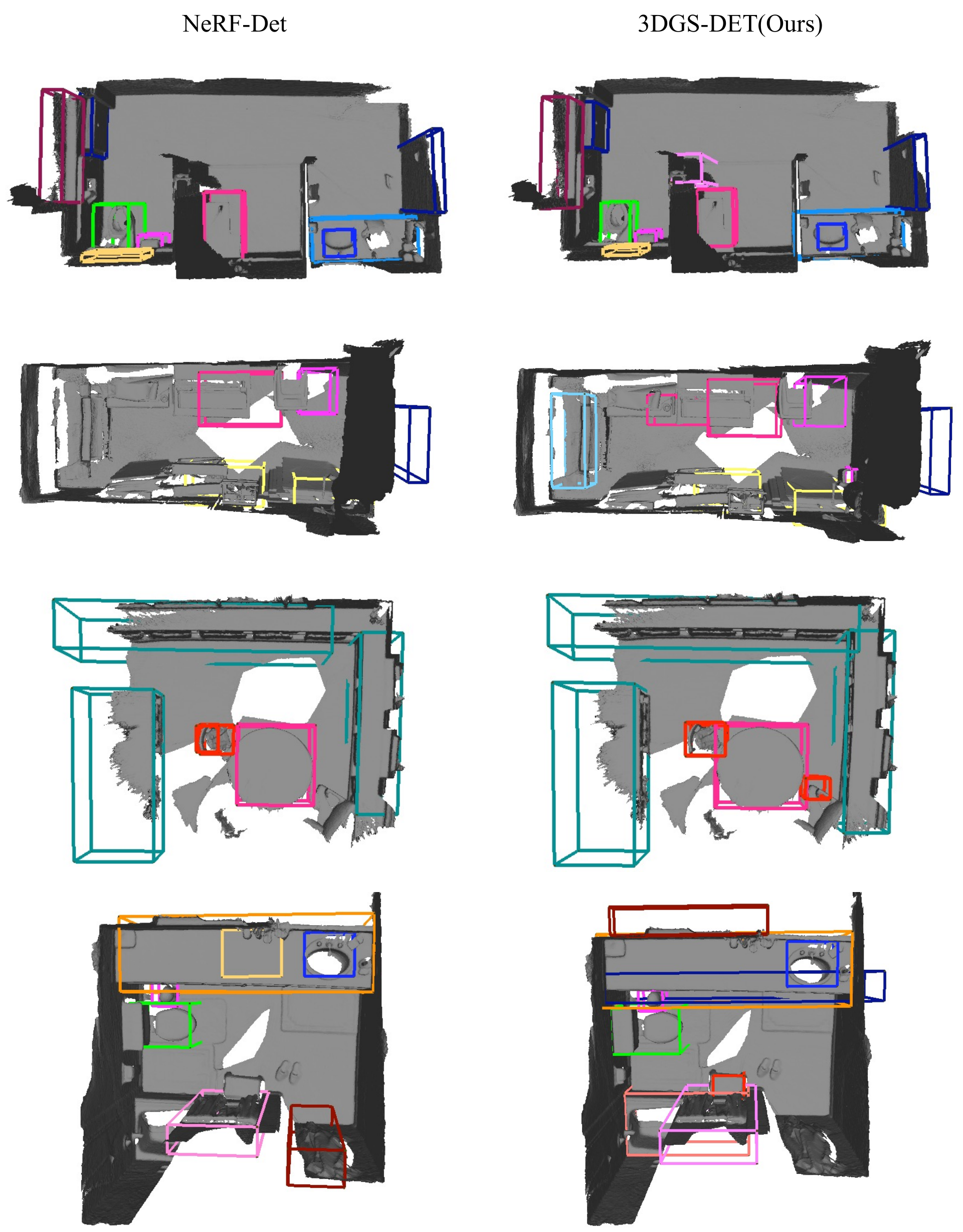}
    \end{overpic}
    \caption{More qualitative comparisons.  Our method detects more objects  with better positional precision, highlighting the advantages of our approach over NeRF-Det~\citep{xu2023nerf}. In this figure, the scene is represented using mesh to clearly display the boxes. Note that black and white boxes indicate predictions with incorrect categories, while boxes of other colors represent predictions with the correct categories.}

    \label{fig:supp_vis_1}

\end{figure*}

\begin{figure*}[ht]
     \centering
    \begin{overpic}[width=0.9\textwidth]
    {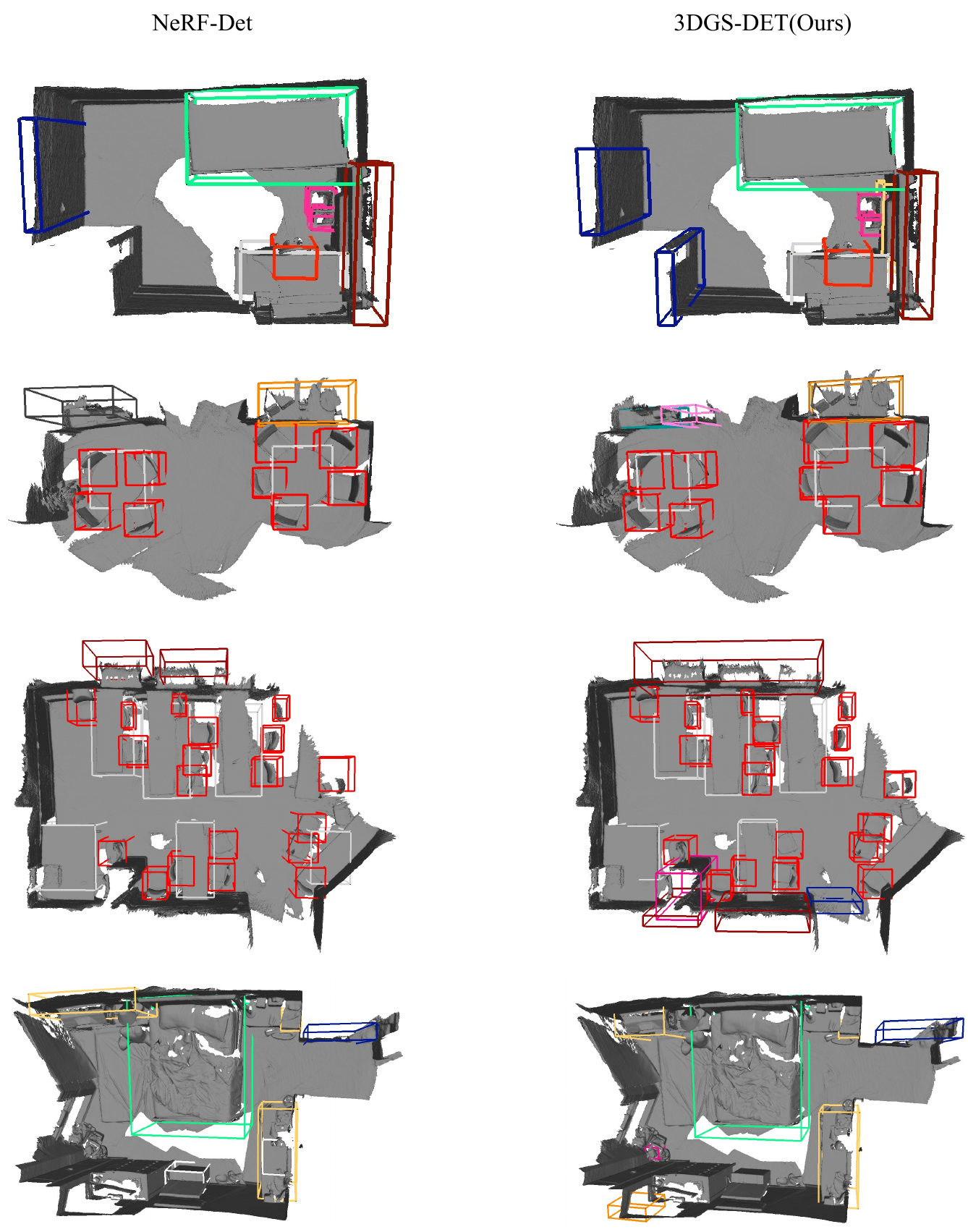}
    \end{overpic}
    \caption{More qualitative comparisons.  Our method detects more objects  with better positional precision, highlighting the advantages of our approach over NeRF-Det~\citep{xu2023nerf}. In this figure, the scene is represented using mesh to clearly display the boxes. Note that black and white boxes indicate predictions with incorrect categories, while boxes of other colors represent predictions with the correct categories.}

    \label{fig:supp_vis_2}

\end{figure*}

\begin{figure*}[h!]
     \centering
    \begin{overpic}[width=0.9\textwidth]
    {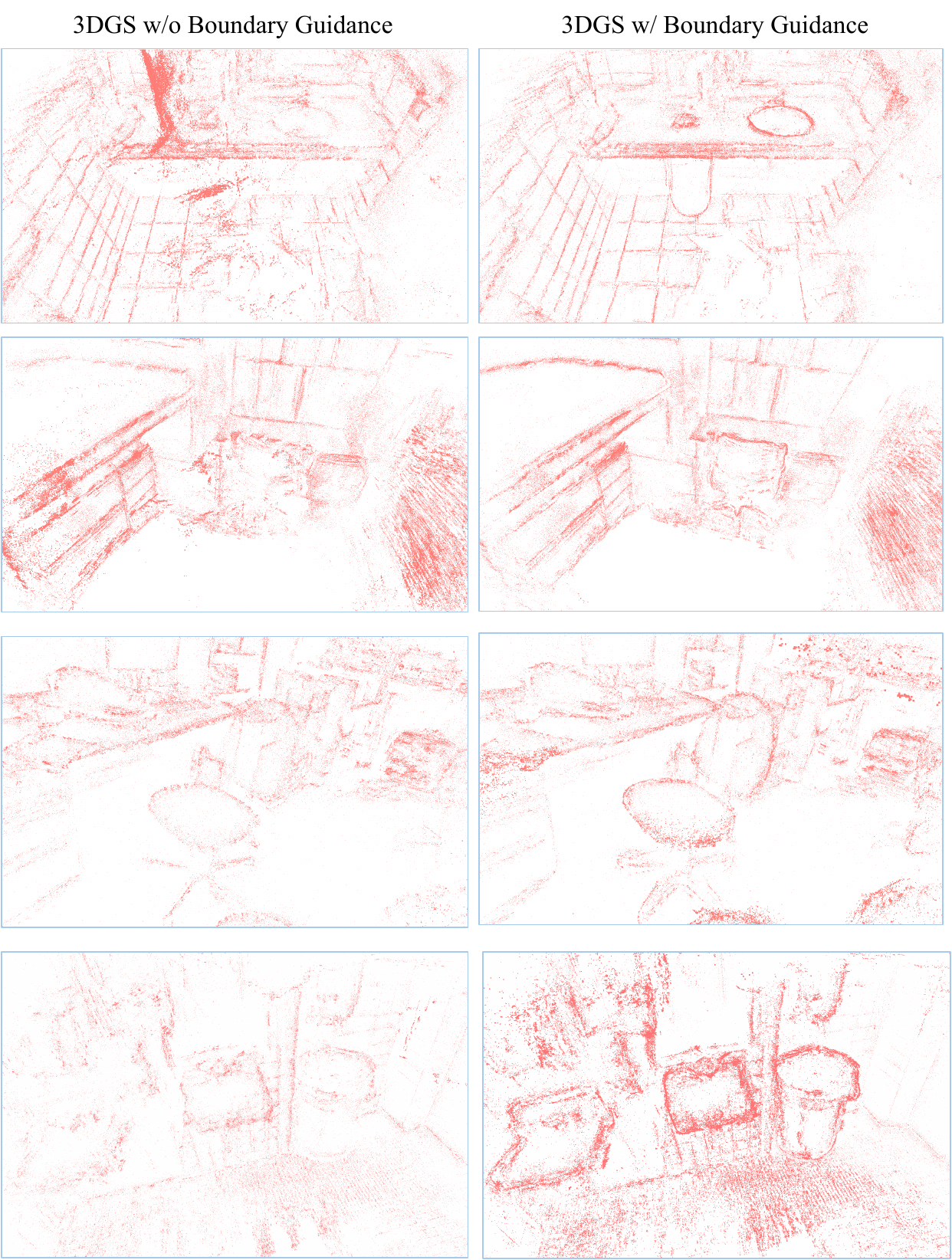}
    \end{overpic}
    \caption{Analysis on the effect of Boundary Guidance. Gaussian blobs trained with Boundary Guidance exhibit clearer spatial distribution and more distinct differentiation between objects and background. Note that we visualize only the positions of the Gaussian blobs to highlight their spatial
distribution, while omitting other attributes for clarity.}

    \label{fig:guidance_supp}

\end{figure*}

\begin{figure*}[h!]
     \centering
    \begin{overpic}[width=0.8\textwidth]
    {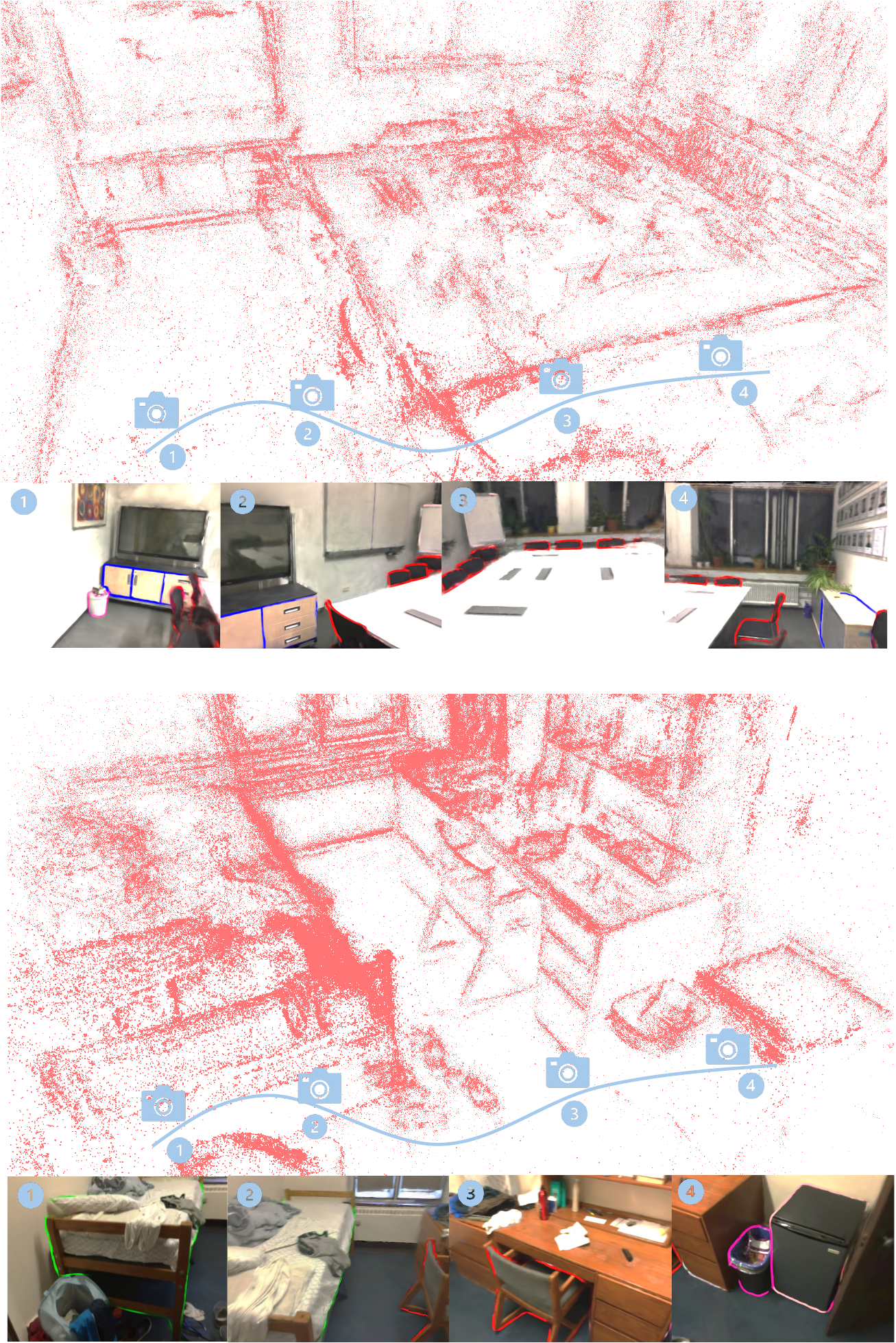}
    \end{overpic}
    \caption{Rendered images from different views by 3DGS trained with Boundary Guidance. The category-specific boundaries are clearly rendered and exhibit multi-view consistency, demonstrating that the 3D representation has successfully embedded the priors provided by Boundary Guidance.}

    \label{fig:render 1}

\end{figure*}

\begin{figure*}[h!]
     \centering
    \begin{overpic}[width=0.825\textwidth]
    {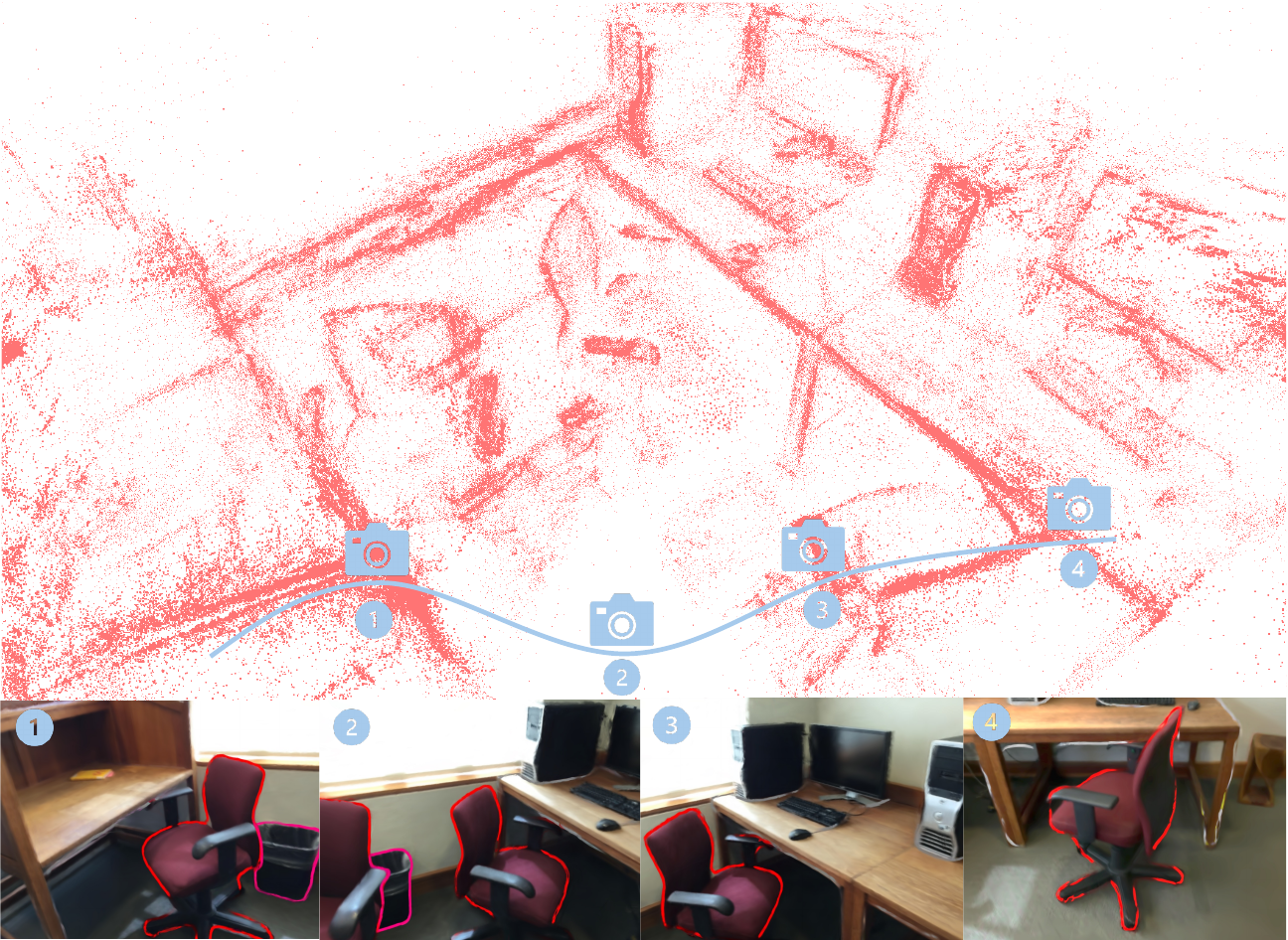}
    \end{overpic}
    \caption{Rendered images from different views by 3DGS trained with Boundary Guidance. The category-specific boundaries are clearly rendered and exhibit multi-view consistency, demonstrating that the 3D representation has successfully embedded the priors provided by Boundary Guidance.}

    \label{fig:render 2}

\end{figure*}

\begin{figure*}[h!]
     \centering
    \begin{overpic}[width=0.98\textwidth]
    {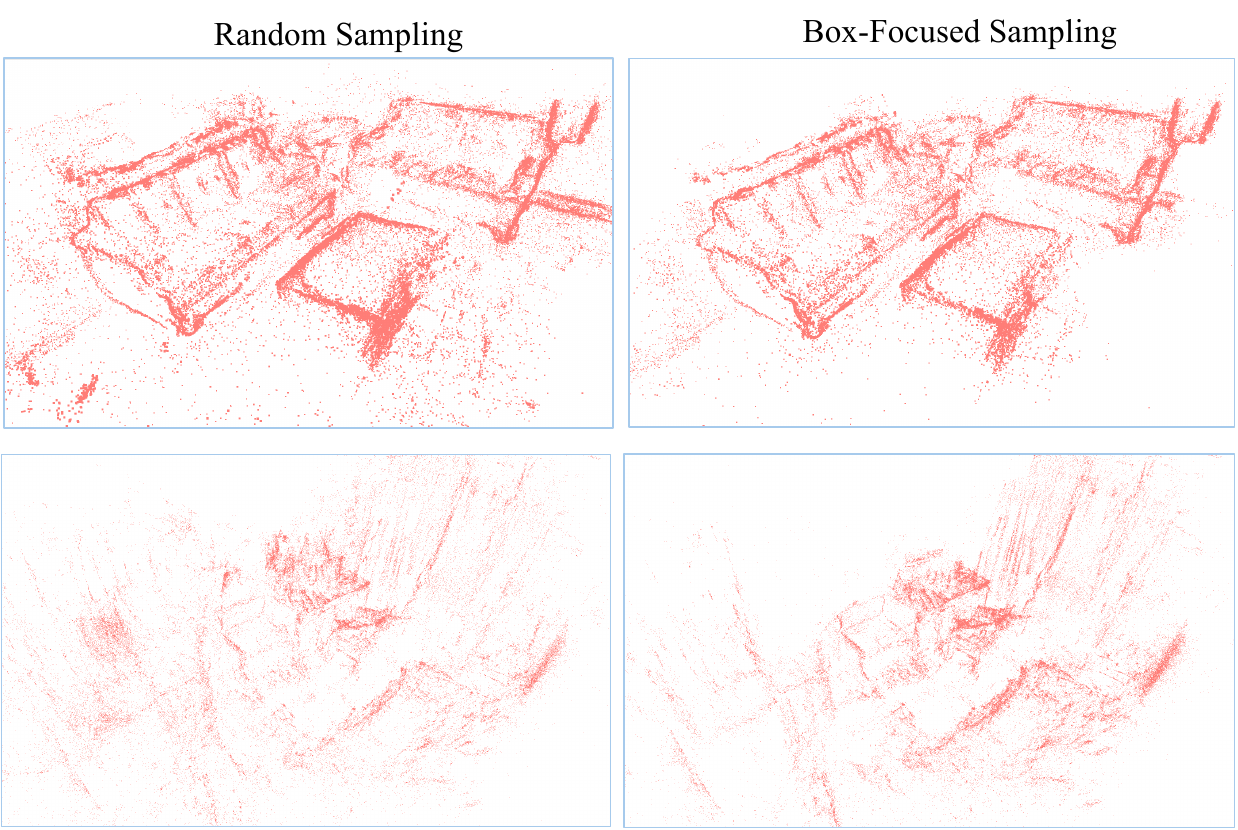}
    \end{overpic}
    \caption{Qualitative Analysis of Box-Focused Sampling for 3DGS Blobs. Box-Focused Sampling significantly retains more object blobs and reduces noisy background blobs. Note that we visualize only the positions of the Gaussian blobs to highlight their spatial distribution, while omitting other attributes for clarity.}

    \label{fig:sample_supp}

\end{figure*}

We provide more qualitative results
in~\figrefn{fig:supp_vis_1} and~\figrefn{fig:supp_vis_2}.
As illustrated, our method detects more objects in the scene with higher positional accuracy than NeRF-Det~\cite{xu2023nerf}, demonstrating the advantages of our method. Note that NeRF-Det++~\cite{nerfdetpp} had not released pretrained models, code, or qualitative results on the official GitHub page by the time of our submission. Therefore, we could not provide a qualitative comparison with NeRF-Det++ here.

\section{Failure Cases}

\begin{figure*}[h!]
     \centering
    \begin{overpic}[width=0.98\textwidth]
    {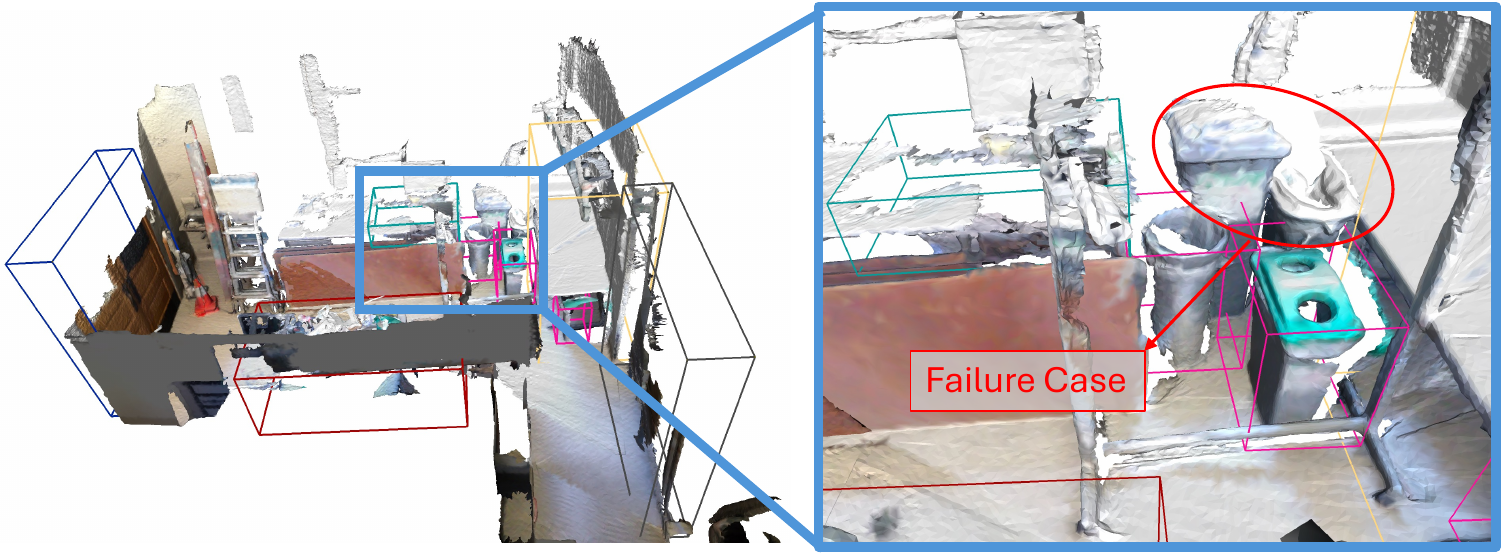}
    \end{overpic}
    \caption{Failure cases. In scenes with severe occlusions, our method exhibits a small number of missed detections. These failure cases are caused by occlusions, which introduce inherent ambiguity in multi-view detection when key visual evidence is obscured.}

    \label{fig:failure_case}

\end{figure*}

We provide failure cases in~\figrefn{fig:failure_case}.
In scenes with severe occlusions, our method exhibits a small number of missed detections. These missed detections are caused by occlusions, which introduce inherent ambiguity in multi-view detection when key visual evidence is obscured.

\section{Limitation and Future Work}
While 3DGS-DET achieves significant improvements over
strong alternative methods, several limitations remain for further exploration.
As the first work to introduce 3DGS into indoor 3D Object Detection~(3DOD), this work mainly focuses on the primary stage of this pipeline: empowering 3DGS for 3DOD. 
Extensive experiments demonstrate that our designs can lead to significant improvements. 
Beyond empowering the 3DGS representation, a subsequent detector specifically designed for 3DGS could hold promise in the future. 

\clearpage
\newpage

\end{document}